\documentclass[10pt,twocolumn,letterpaper]{article}

\usepackage{cvpr}              

\definecolor{cvprblue}{rgb}{0.21,0.49,0.74}
\usepackage[pagebackref,breaklinks,colorlinks,allcolors=cvprblue]{hyperref}
\usepackage{multirow}
\def\paperID{1802} 
\def\confName{CVPR}
\def\confYear{2025}

\title{Towards Natural Language-Based Document Image Retrieval: \\New Dataset and Benchmark}

\author{Hao Guo$^{1,3,4,}\thanks{Equal contribution. ~~\textsuperscript{\dag}Corresponding authors.}$ , Xugong Qin$^{2,*,\dagger}$, Jun Jie Ou Yang$^5$, Peng Zhang$^{2,6,\dagger}$, Gangyan Zeng$^2$, \\Yubo Li$^{1,3,4}$, Hailun Lin$^{1,3}$\\
$^1$Institute of Information Engineering, Chinese Academy of Sciences\\
$^2$School of Cyber Science and Engineering, Nanjing University of Science and Technology\\
$^3$State Key Laboratory of Cyberspace Security Defense\\
$^4$School of Cyber Security, University of Chinese Academy of Sciences\\
$^5$University of Southern California, $^6$Laboratory for Advanced Computing and Intelligence Engineering\\
{\tt\small guohao2022@iie.ac.cn, qinxugong@njust.edu.cn}
}

\begin{document}
\maketitle
\begin{abstract}
Document image retrieval (DIR) aims to retrieve document images from a gallery according to a given query. Existing DIR methods are primarily based on image queries that retrieve documents within the same coarse semantic category, e.g., newspapers or receipts. However, these methods struggle to effectively retrieve document images in real-world scenarios where textual queries with fine-grained semantics are usually provided. To bridge this gap, we introduce a new Natural Language-based Document Image Retrieval (NL-DIR) benchmark with corresponding evaluation metrics. In this work, natural language descriptions serve as semantically rich queries for the DIR task. The NL-DIR dataset contains 41K authentic document images, each paired with five high-quality, fine-grained semantic queries generated and evaluated through large language models in conjunction with manual verification. We perform zero-shot and fine-tuning evaluations of existing mainstream contrastive vision-language models and OCR-free visual document understanding (VDU) models. A two-stage retrieval method is further investigated for performance improvement while achieving both time and space efficiency. We hope the proposed NL-DIR benchmark can bring new opportunities and facilitate research for the VDU community. Datasets and codes will be publicly available at \url{huggingface.co/datasets/nianbing/NL-DIR}.
\end{abstract}    
\vspace{-5pt}
\section{Introduction}
\label{sec:intro}

With the rapid development of mobile cameras and smartphones, document images become one of the most convenient ways to record and disseminate information. As an important part of information retrieval (IR), document image retrieval (DIR) aims to accurately and efficiently retrieve relevant document images from large repositories based on user queries. 

\begin{figure}[t]
\begin{center}
\includegraphics[width=\linewidth]{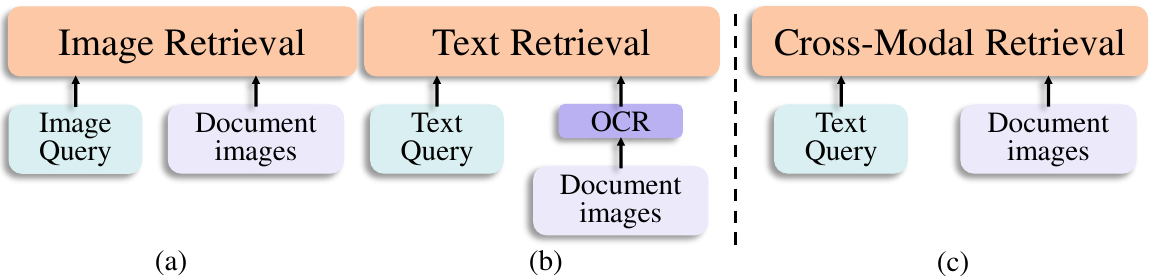}
\end{center}
\vspace{-20px}
\caption{Comparison of different paradigms for DIR tasks. (a): Query-by-image. (b): Query-by-text. (c): Our approach directly retrieves document images using natural language queries.}
\vspace{-20px}
\label{fig:difretrieval_method}
\end{figure}

As illustrated in \cref{fig:difretrieval_method}, considering the type of queries, existing DIR methods can be categorized into two paradigms: query-by-image and query-by-text. Specifically, query-by-image methods \cite{lu2004documentdatabase,liu2008mobile,alaei2016brief,alaei2019comparative,dixit2021document,liu2023end,borenstein2023phd} directly extract features from query and document images for matching. The extraction of image features primarily considers visual information, which limits DIR within the same coarse semantic class, e.g., newspaper or receipt. Comparatively, query-by-text \cite{karpukhin2020dense,xiong2020approximate,zhan2021optimizing} resorts to Optical Character Recognition (OCR) engines \cite{qin2021mask,qin2023towards,qiao2020seed,qiao2021pimnet,zeng2024fdp} to extract words from document images. The retrieval models mainly focus on enhancing the textual representations of queries and documents, while commonly struggling to capture the extensive and comprehensive visual features presented in documents, such as visual elements, display styles, content layout, etc.

To summarize, the current retrieval methods are restricted to single-modality matching, which inevitably results in information loss and noise for the original text or image domain. Consequently, it is necessary to design a cross-modal retrieval model for DIR that can handle text queries and visual documents simultaneously (see \cref{fig:difretrieval_method} (c)). This new DIR task is non-trivial as it involves multiple research areas, including text understanding, document image understanding, and cross-modal learning, and is also a crucial component of various downstream tasks.

An intuitive solution is to directly apply existing cross-modal retrieval methods \cite{radford2021clip,li2022blip,bao2022vlmo,li2023blip2,yu2022coca} on this task. However, most of these methods are dedicated to modeling the correspondence between textual descriptions and natural images, rather than document images. The lack of a dedicated, fine-grained DIR dataset prevents existing models from obtaining sufficient supervised signals, thereby limiting their performance on DIR tasks. It is worth noting that, concurrently with our work, recent studies have utilized web screenshots \cite{ma2024unifying} or PDF images \cite{faysse2024colpali} to construct the training data. Unfortunately, these studies have not been adequately evaluated on relatively large-scale document images from real-world scenarios.

In this work, we propose the Natural Language-based DIR (NL-DIR) dataset to train and evaluate models' capabilities on cross-modal retrieval in the document domain. NL-DIR is composed of 41K document images with 205K queries, featuring a diverse collection of real-world document images from the Industry Documents Library\footnote{\url{https://www.industrydocuments.ucsf.edu}}. 10\% of data in NL-DIR is allocated for evaluation to construct the benchmark. We employ layout-aware methods \cite{wang2023layout_aware,lamott2024lapdoc} and leverage large language model (LLM) to generate image-query pairs. After scoring with various models, manual verification is performed for filtering.

To evaluate the existing mainstream models, we propose a two-stage approach including a recall and a re-ranking stages. Specifically, in the recall stage, contrastive visual-language models or generative VDU models are employed for the retrieval of the top 100 results from a substantial corpus of documents. In the re-ranking stage, a cross-attention module is incorporated to reorder the top 100 results, thus yielding the final DIR results. The recall and mean reciprocal rank are taken as the main evaluation metrics. Finally, we compare the proposed NL-DIR model with the OCR-dependent text document retrieval models and existing large vision-language model (LVLM) based DIR methods considering both accuracy and efficiency.

The key insights from the study include: 
\begin{itemize}[itemsep=0pt,topsep=2pt,leftmargin=20pt] 
    \item The selection of pre-training tasks and datasets significantly impacts the retrieval performance. Models pre-trained on image-text contrastive learning tasks demonstrate superior performance.
    \item Fine-grained interaction is quite an effective way to capture semantic information in document images.
    \item OCR-free models exhibit advantages when queries contain visual information of non-text elements, particularly in low-quality document images.
\end{itemize}

The contributions of this work can be summarized as follows: 1. Providing the first \textbf{benchmark} for fine-grained DIR in natural scenes by releasing a publicly available dataset consisting of 41K document images with 205K queries; 2. Conducting a comprehensive \textbf{analysis} on the performance of popular cross-modal retrieval and document understanding models; 3. Proposing a two-stage \textbf{method} for DIR in real-world scenarios that achieves strong retrieval performance while ensuring efficient use of time and space throughout the retrieval process.
\vspace{-5pt}
\section{Related Work}
\label{sec:Related_Work}

\label{headings}
\vspace{-3pt}
\subsection{Document Image Retrieval}
\vspace{-3pt}

Traditional content-based DIR methods \cite{lu2004documentdatabase,liu2008mobile,alaei2016brief,alaei2019comparative,dixit2021document,liu2023end,borenstein2023phd} achieve retrieval by extracting features and computing similarities between the query image and the document images. This matching process can be performed either at the global level or at the single-word level. Besides, common cross-modal retrieval methods \cite{radford2021clip,li2022blip,li2021albef,bao2022vlmo,yu2022coca,chen2024internvl} perform retrieval in natural scenes through techniques like image-text matching. Motivated by this, Wang et al. \cite{wang2021tdsl} achieve accurate scene text retrieval through matching proposals and query words. However, the queries within these methods are represented for simple classes or single words with coarse semantics. DIR based on semantic-rich queries remains unexplored.

Among recent DIR methods, PHD \cite{borenstein2023phd} explores an image-to-image retrieval method for historical documents. TransferDoc \cite{bakkali2023transferdoc} adopts document category retrieval on existing datasets with classification labels \cite{harley2015rvlcdip}. However, these methods do not effectively utilize the information from image and text modalities. VILE \cite{yuan2023vile}  generates web page images from existing retrieval datasets \cite{craswell2021MARCOV2} and utilizes these images as auxiliary information for text retrieval. Concurrent studies \cite{ma2024unifying,faysse2024colpali} employ document screenshots, such as those from PDFs or Wikipedia pages, and utilize large visual-language encoders for alignment modeling. Nevertheless, these approaches are difficult to generalize to DIR in large-scale real-world scenarios.

\vspace{-3pt}
\subsection{Visual Document Understanding}
\vspace{-3pt}

Robust representation of document images matters a lot for the DIR tasks. Recent research primarily focuses on two distinct approaches: OCR-dependent and OCR-free models. In OCR-dependent document understanding models \cite{xu2020layoutlm,xu2020layoutlmv2,huang2022layoutlmv3,appalaraju2021docformer,appalaraju2023docformerv2,yu2023structextv2,peng2022ernielayout,li2021selfdoc,tang2023udop}, image, text, and layout information are integrated as inputs. By optimizing pre-training tasks and model structures, these models have achieved significant performance improvements in common downstream document understanding tasks. OCR-free models \cite{kim2022donut,davis2022dessurt,li2022dit,lee2023pix2struct,blecher2023nougat} directly model image pixels, thereby avoiding the accumulation of OCR errors and exhibiting greater robustness. Donut \cite{kim2022donut} designs the text reading task to output continuous text sequences. Pix2Struct \cite{lee2023pix2struct} designs a screenshot parsing task to generate the HTML DOM tree for webpage screenshots. Nougat \cite{blecher2023nougat} is trained to parse the structured representation of academic documents directly from PDF images.

In parallel, existing vision-language models (VLMs) \cite{lv2023kosmos2.5,zhang2023llavar,ye2023mplugDOCOWL,feng2023docpedia,ye2023ureader,hu2024mplug1.5,liu2024textmonkey,hu2024mplugdocowl2} utilize datasets such as charts, tables, and documents to improve the abilities on document understanding with the improved document image reading designs. For instance, Vary \cite{wei2023vary} introduces an additional image encoder. UReader \cite{ye2023ureader} employs a shape-adaptive cropping module to divide the original image into multiple sub-images. DocOwl1.5 \cite{hu2024mplug1.5} utilizes an H-Reducer, which convolves horizontally adjacent patches to effectively comprehend high-resolution images. TextMonkey \cite{liu2024textmonkey} incorporates shifted window attention and employs a resampler to filter important tokens. Above models are capable of creating more comprehensive and accurate document representations and performing well in tasks such as information extraction, visual question answering, document classification, and layout analysis. Nevertheless, the efficacy of document understanding models in the foundational task of NL-DIR remains to be investigated.

\vspace{-3pt}
\subsection{Construction of Document Image Datasets}
\vspace{-3pt}

For construction of NL-DIR, we consider existing datasets on text document retrieval \cite{craswell2021MARCOV2,thakur2021beir} and document understanding \cite{harley2015rvlcdip,svetlichnaya2020Deepform,mathew2021docvqa,pfitzmann2022doclaynet,van2023DUDE,ocridl2022}. Between the categories, document image datasets cover a wide variety of image types (letters, forms, receipts, etc.) and tasks (document classification, key information extraction, question answering, and document layout analysis, etc.), which provide a solid foundation for VDU tasks. In recent years, researchers have proposed several innovative approaches to construct document image datasets. Pix2Struct \cite{lee2023pix2struct} and VILE \cite{yuan2023vile} utilize existing document datasets and web corpus to crawl and collect web page screenshots, while other work \cite{kim2022donut,borenstein2023phd} focuses on document rendering generation. 

The success of large language models has also inspired recent research \cite{wang2023layout_aware,liu2023MMC,li2023Monkey,nayak2024Bonito,chen2024allava} exploring their use for data annotation. Typically, MMC \cite{liu2023MMC} utilizes GPT-4 \cite{achiam2023gpt4} to generate Q\&A pairs associated with diagrams to build instruction data for a variety of diagram comprehension tasks. Monkey \cite{li2023Monkey} uses ChatGPT \cite{john2023chatgpt} to generate detailed and rich image description data. Following the above methods, to construct the NL-DIR dataset, we resort to the use of LLM for efficient data generation and filtering.
\vspace{-5pt}
\section{NL-DIR: Natural Language-Based Document Image Retrieval}

\vspace{-3pt}
\subsection{Dataset Overview}
\vspace{-3pt}

The NL-DIR aims to build a dataset of DIR with fine-grained semantic descriptions, which is constructed following two main stages: data collection and data annotation. The introduced fine-grained description is a relative concept compared to coarse-grained semantics. Unlike traditional DIR methods that typically rely on categories of document images, NL-DIR retrieval documents are based on queries with more concrete descriptions through natural languages. These queries are constructed based on the document content and can include specific details in documents.


Overall, NL-DIR consists of 41,795 document images with a diverse range of document types, and each image corresponds to five high-quality fine-grained semantic queries. Following an 8:1:1 ratio partition, the dataset is divided into three sets: training, validation, and testing, with each set maintaining an identical distribution of document categories. The test set is employed to construct a benchmark to evaluate the cross-modal retrieval capabilities of various models on document images. Additionally, we will also release the multi-class labels and multi-page OCR results in NL-DIR to support document classification, multi-page retrieval, and other tasks.

\vspace{-5px}
\begin{figure}[!htb]
\centering
\begin{subfigure}{0.32\linewidth}
    \centering
    \includegraphics[width=\linewidth]{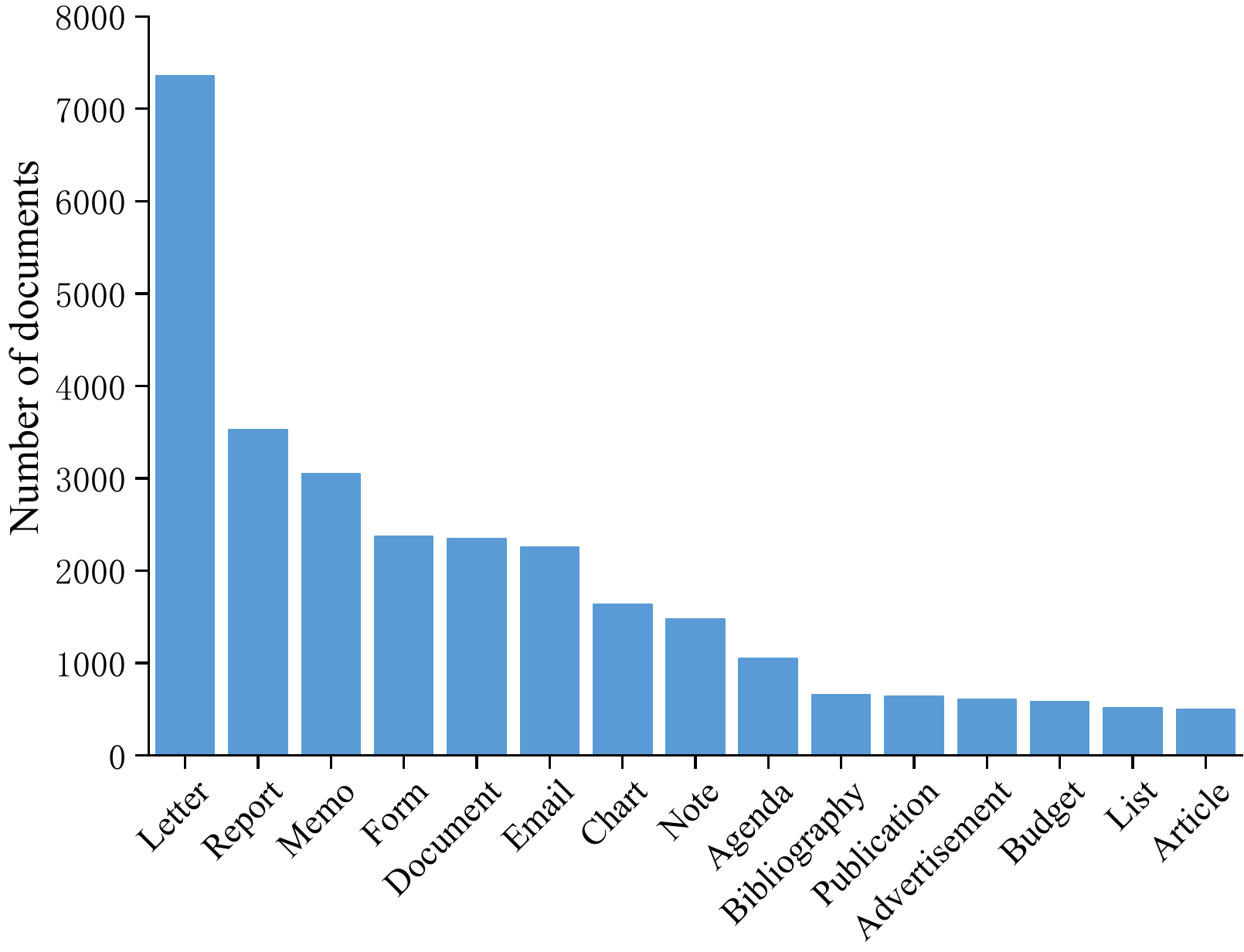}
    \caption{}
    \label{subfig:doctypes}
\end{subfigure}
\hfill
\begin{subfigure}{0.32\linewidth}
    \centering
    \includegraphics[width=\linewidth]{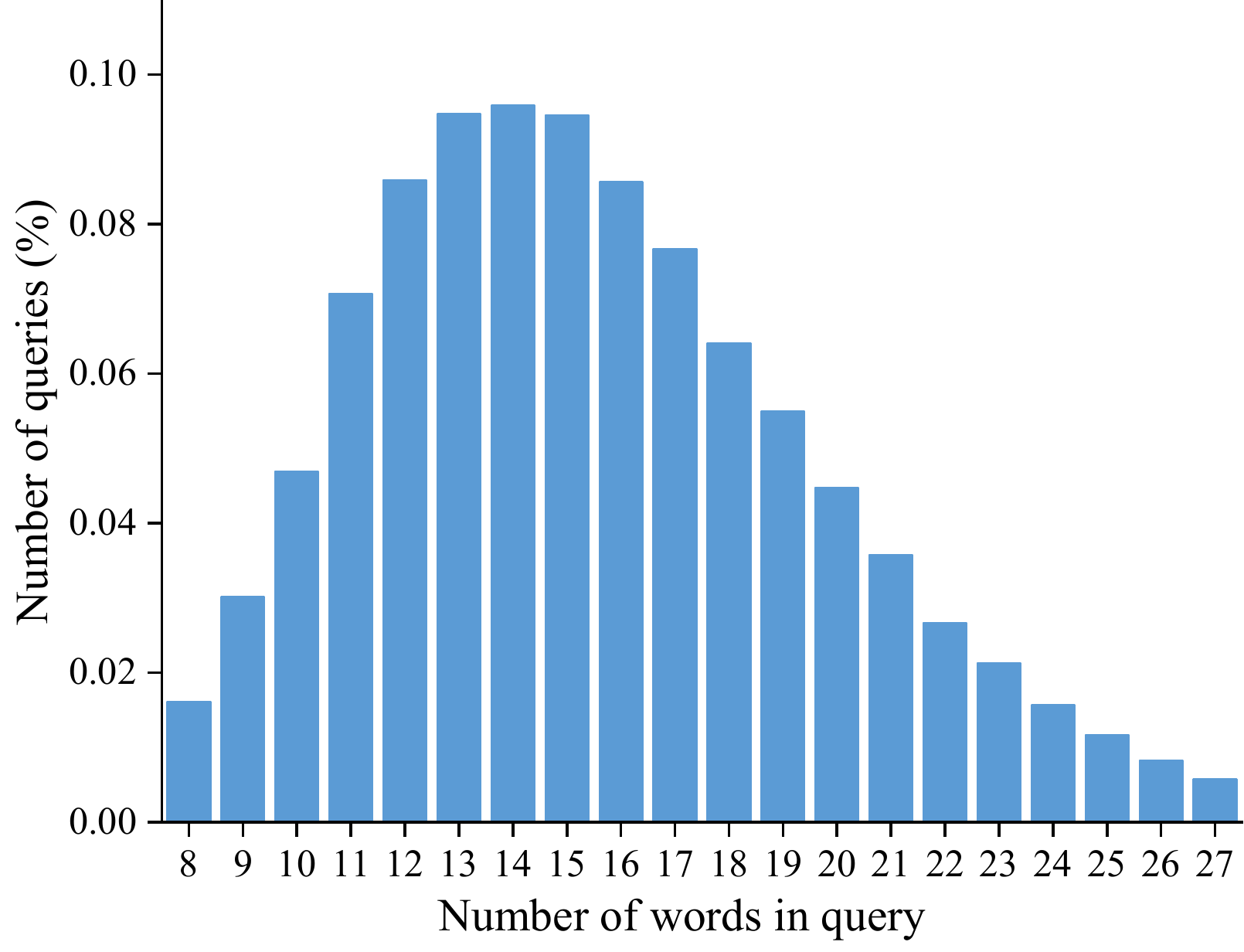}
    \caption{}
    \label{subfig:query_nums_words}
\end{subfigure}
\hfill
\begin{subfigure}{0.32\linewidth}
    \centering
    \includegraphics[width=\linewidth]{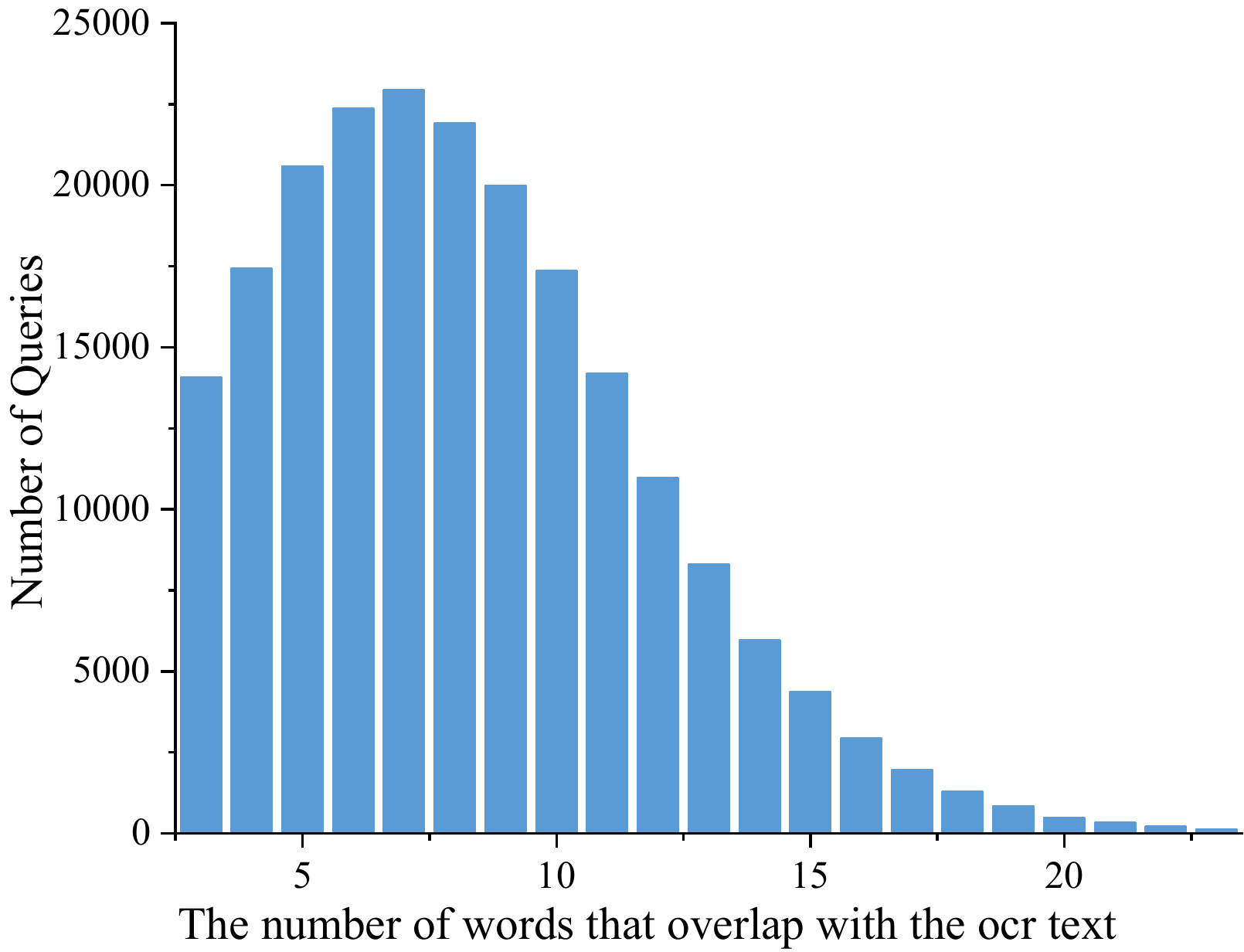}
    \caption{}
    \label{subfig:query_overlap}
\end{subfigure}
\vspace{-10px}
\caption{Statistics of NL-DIR. (a) Various types of documents. (b) Queries with a particular length. (c) Query and OCR text overlap situation. Best zoom to view.}
\label{fig:dataset_statistic}
\end{figure}
\vspace{-10px}

\vspace{-10px}
\begin{figure}[!htb]
\begin{center}
    \includegraphics[width=0.85\linewidth]{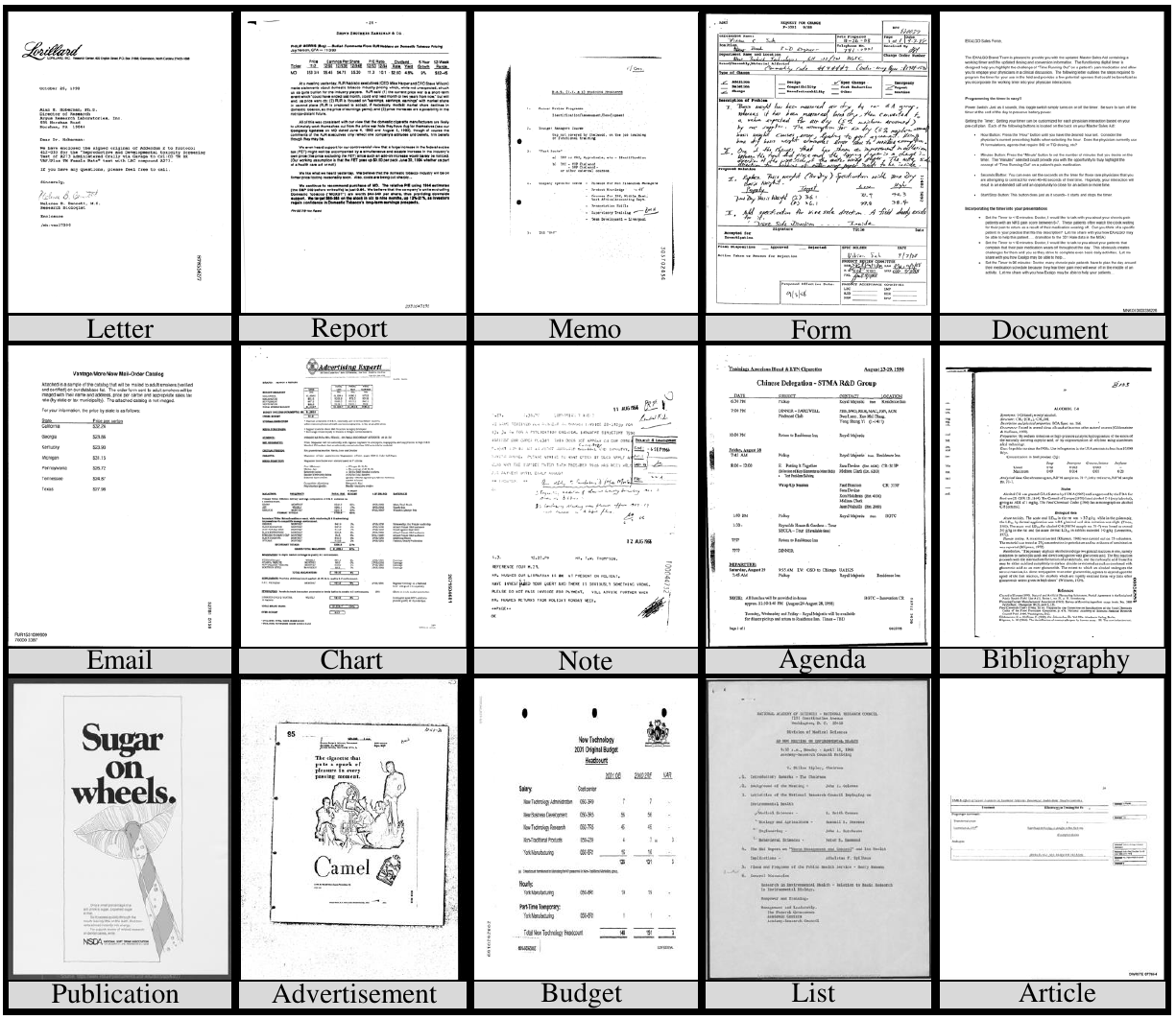}
\end{center}
\vspace{-20px}
\caption{Examples of various types of document images.}
\label{fig:examples}
\end{figure}
\vspace{-10px}

The NL-DIR dataset contains documents of 247 categories, and we perform statistical analysis on the top 15 document categories, with the results presented in \cref{subfig:doctypes}. We randomly select some document image examples from the top 15 categories and include them in \cref{fig:examples}. \cref{subfig:query_nums_words} illustrates that the distribution of query lengths is primarily concentrated between 10 and 20 words. The shortest query contains at least eight words, ensuring the quality of fine-grained semantic description. \cref{subfig:query_overlap} shows the number of overlapping words between the queries and the original OCR text. At least three or more words in the queries appear in the original OCR text,  with the majority of queries containing approximately seven overlapping words. Although some of these overlapping words may be stop words, they still provide a degree of assurance regarding the validity of the generated queries.

\vspace{-10px}
\begin{figure}[!htb]
\begin{subfigure}{0.46\linewidth}
    \centering
    \includegraphics[width=\linewidth]{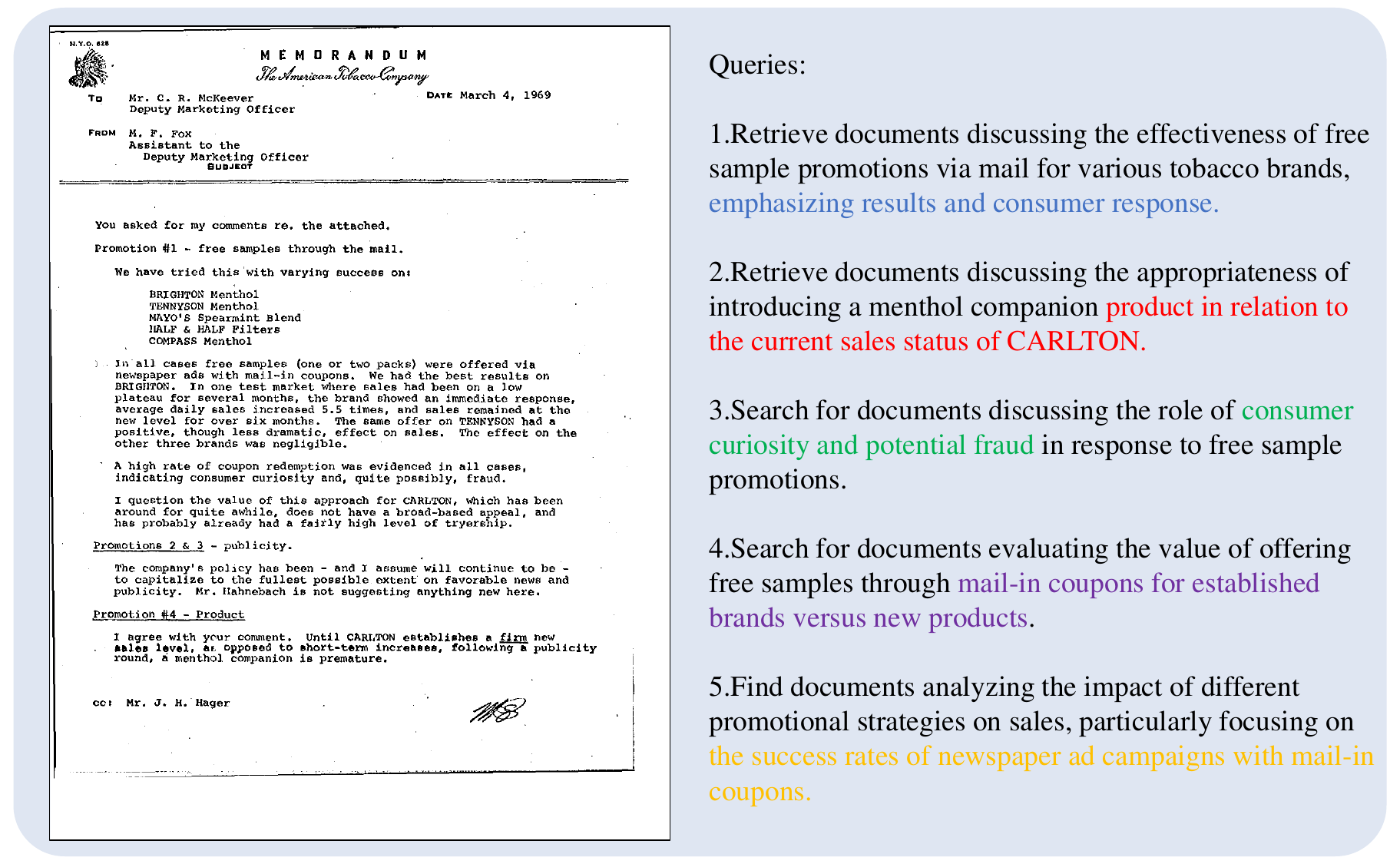}
    \caption{}
    \label{subfig:letter}
\end{subfigure}
\hfill
\begin{subfigure}{0.46\linewidth}
    \centering
    \includegraphics[width=\linewidth]{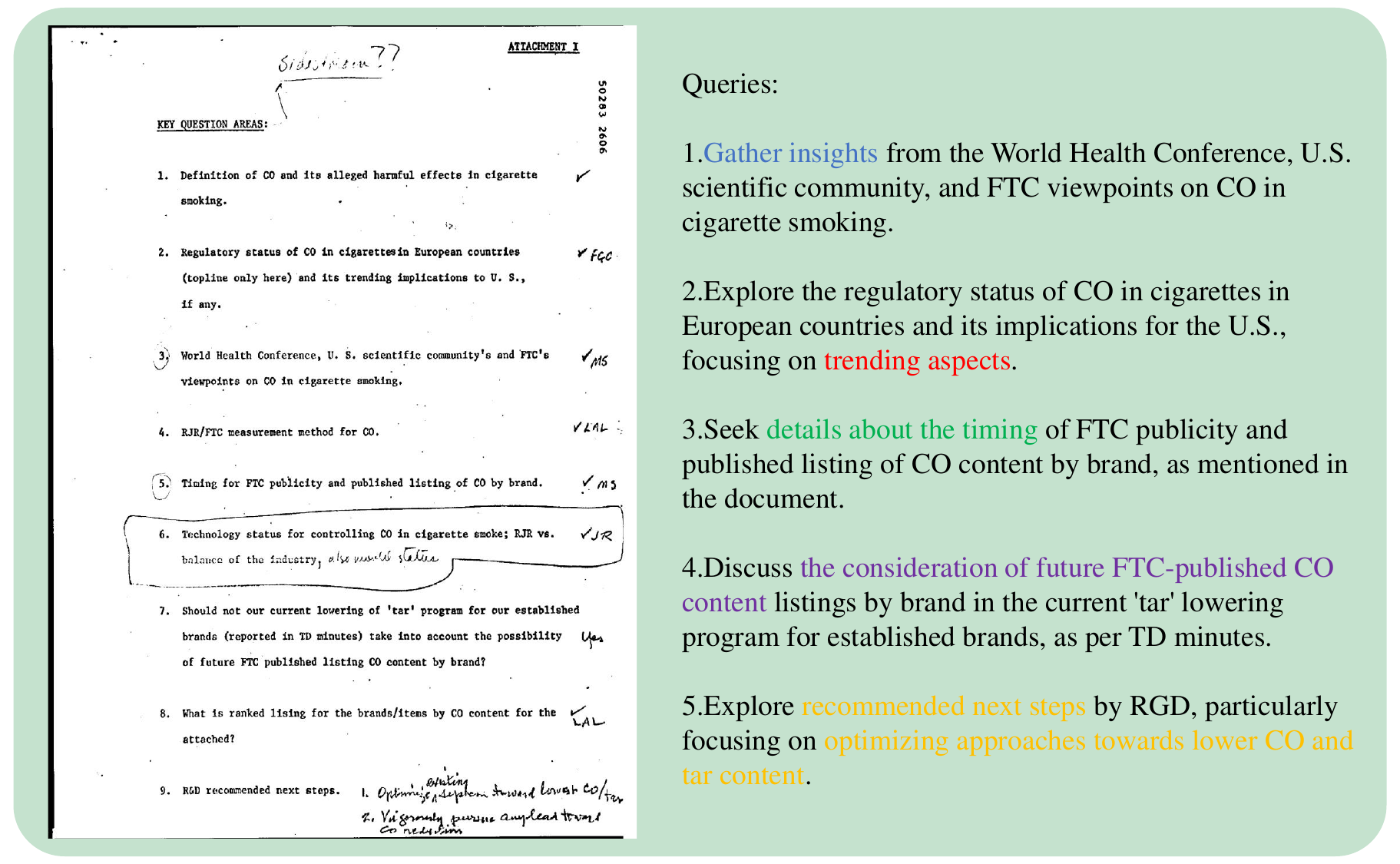}
    \caption{}
    \label{subfig:report}
\end{subfigure}
\vspace{-10px}
\caption{Examples of queries associated with different types of document images. Best zoom to view.}
\label{fig:letter_report}
\end{figure}
\vspace{-10px}

To provide a direct illustration of the diversity in document image types and their corresponding queries, we present two types of common document images along with related example analyses. As shown in \cref{fig:letter_report}, the vocabulary associated with document images is highlighted in different colors in each query to emphasize the variety of generated queries. When the textual information is abundant and structured information is limited in \cref{fig:letter_report}(a), the five corresponding queries primarily focus on comprehending the overall content of the image. In \cref{fig:letter_report}(b), when the document image contains specific entries, the queries tend to be more granular, with distinct focal points for each entry.



\vspace{-3pt}
\subsection{Data Collection}
\vspace{-3pt}

To construct NL-DIR, we first collect document images from real-world scenarios that feature well-structured layouts and rich textual content, including letters, reports, forms, documents, etc. Specifically, we gather approximately 50k document images from OCR-IDL \cite{ocridl2022} and supplement them with document images from DocVQA \cite{mathew2021docvqa}. After removing duplicates, we obtain a dataset comprising 60k initial documents.

These document images are sourced from the Industry Documents Library, a digital archive of files created by industries that affect public health, hosted by the University of California, San Francisco Library. The corresponding layout text information is extracted using Microsoft OCR for DocVQA and Amazon Textract for OCR-IDL. Compared to open-source OCR engines, these OCR annotations have better quality. The collected images and layout text information form the foundation for dataset construction.

\vspace{-3pt}
\subsection{Data Annotation}
\vspace{-3pt}

\begin{figure}[t]
\begin{center}
    \includegraphics[width=0.95\linewidth]{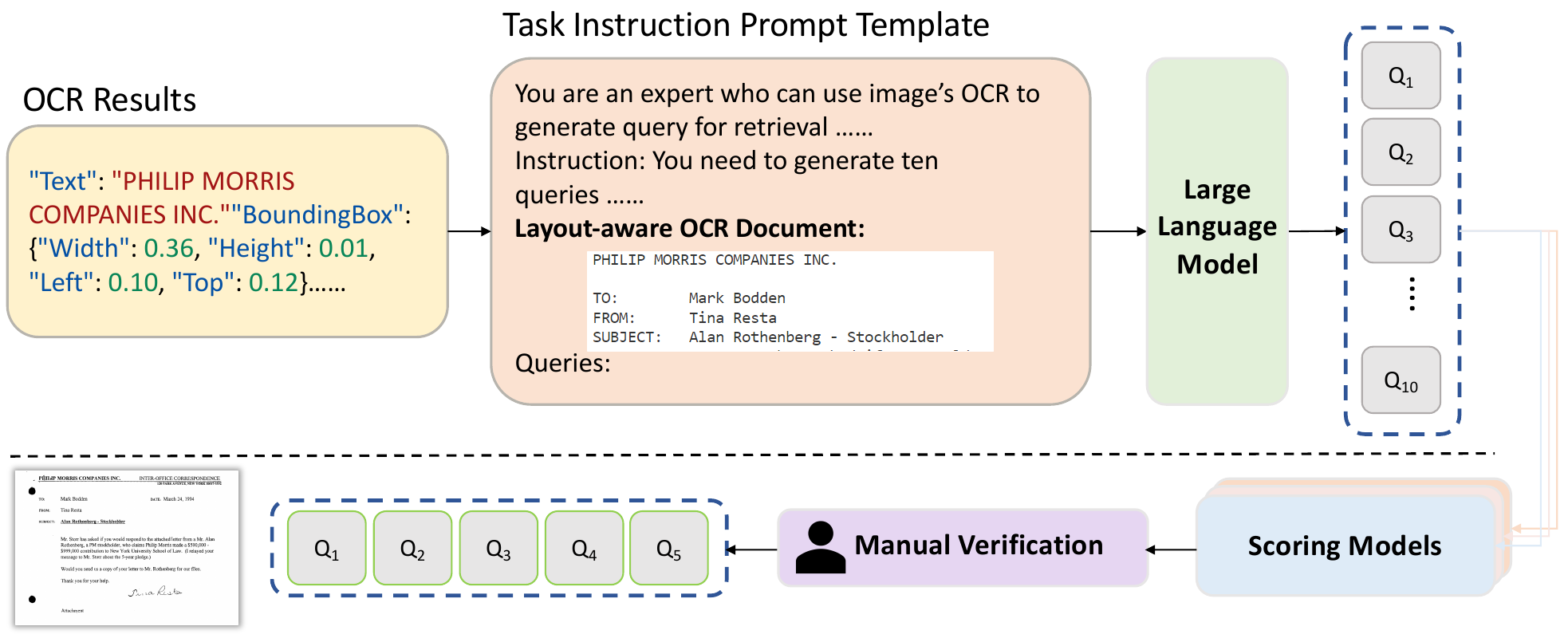}
\end{center}
\vspace{-20px}
\caption{The pipeline for query generation and filtering.}
\label{fig:generatequery}
\vspace{-15px}
\end{figure}

Given the real-world document images, the next step is to generate high-quality queries matching the documents. To achieve this, 
we propose a pipeline that involves the generation, filtering, and verification processes as shown in \cref{fig:generatequery}.

Due to the low efficiency and difficulty in ensuring the quality of manually constructed queries, and the limited VDU ability of existing LVLMs, we plan to leverage the powerful text understanding and generation capabilities of ChatGPT to achieve high-quality query generation. Specifically, the following two steps are taken: 
\begin{enumerate}[itemsep=0pt,topsep=2pt,leftmargin=20pt] 
    \item To generate the layout-aware document, we utilize the OCR results of obtained document images, employing spaces and line breaks to simulate layout information\footnote{Several related works \cite{wang2023layout_aware,lamott2024lapdoc} have demonstrated that utilizing implicit layout information can help text-only models better understand documents. For this reason, in our work, we leverage layout text information to guide LLMs in generating higher-quality queries.}.
    \item We leverage large language models to generate appropriate queries with the guidance of prompts and simulated layout information.
\end{enumerate}

Although the design of prompt and layout-aware document text can promote the effectiveness of query generation, the generated content still suffers from hallucinations and meaningless results. Therefore, we utilize the following models for further filtering:

\begin{enumerate}[itemsep=0pt,topsep=2pt,leftmargin=20pt] 
    \item Using ChatGPT \cite{john2023chatgpt} and Qwen-VL-Plus \cite{bai2023qwen} to score the generated queries. In particular, we also prompt these models to provide rationales for interpreting the score to prevent bias during scoring.
    \item Generating approximate matching scores for each image-query pair using image-text alignment models, specifically, CLIP \cite{radford2021clip} and BLIP \cite{li2022blip}.
    \item Leveraging the scores generated in the preceding steps, we weight each model's score with 3:3:2:2, sort them in descending order, and visualize them for experienced researchers for manual verification. Additionally, we impose filtering rules including: removing low-quality images; and filtering out queries with low image relevance, strong layout dependence, and excessive generalization.
\end{enumerate}


\vspace{-5pt}
\section{Methodology}

\vspace{-3pt}
\subsection{Retrieval Baselines} 
\vspace{-3pt}

Upon the constructed new NL-DIR dataset and task, we evaluate a wide range of baseline models in terms of retrieval performance and visual representation ability in the document domain. The utilized baseline models are summarized in \cref{tab:baseline_models}, including existing mainstream contrastive VLMs and generative VDU models.

\vspace{-10px}
\paragraph{Contrastive VLMs.} Contrastive VLMs are often equipped with a two-tower architecture that includes a visual encoder and a text encoder, employing contrastive loss to align the representations of textual content with their corresponding visual content representations, such as CLIP \cite{radford2021clip}, BLIP \cite{li2022blip}, SigLIP \cite{zhai2023siglip}, and InternVL-14B-224px \cite{chen2024internvl}. These methods typically perform well in natural scenes but exhibit limited capabilities in the document domain as their visual components are generally not optimized for document scenarios. 

\vspace{-15px}
\paragraph{Generative VDU Models.} Current VDU models commonly use a generative framework that employs decoder-only textual architecture to encode multimodal inputs and generate text. To this end, we evaluate the effectiveness of several generative VDU models \cite{kim2022donut,lee2023pix2struct,blecher2023nougat,wei2023vary,ye2023ureader,liu2024textmonkey,ye2023mplugDOCOWL,wang2024qwen2vl} on NL-DIR, which have explicitly learned the OCR abilities and perform well on downstream VDU tasks. Concurrent works \cite{ma2024unifying,faysse2024colpali} utilize LVLMs to optimize document image retrieval through contrastive learning on web or document datasets, achieving promising results.

\vspace{-10px}
\begin{table}[!htb]
\caption{The statistics of the evaluated baseline models. “FT” denotes fine-tuning with task-specific data.}
\vspace{-10px}
\resizebox{0.48\textwidth}{!}{%
\begin{tabular}{l|c|cc|cc}
\toprule
\multirow{2}{*}{Model} & \multirow{2}{*}{Resolution}       & \multicolumn{2}{c|}{Explicit OCR Learning}                                   & \multicolumn{2}{c}{Parameters}                                                    \\ \cmidrule(lr){3-4}\cmidrule(lr){5-6}
                       &                                   & \multicolumn{1}{c|}{Data Type}                       & Data Size             & \multicolumn{1}{c|}{Visual} & \multicolumn{1}{c}{Textual}                 \\ \midrule
CLIP-base \cite{radford2021clip}  & 224$\times$224                    & \multicolumn{1}{c|}{-}             & -                     & \multicolumn{1}{c|}{87M}    & \multicolumn{1}{c}{63M}                     \\
BLIP-base \cite{li2022blip}       & 224$\times$224                    & \multicolumn{1}{c|}{-}             & -                     & \multicolumn{1}{c|}{86M}    & \multicolumn{1}{c}{137M}                     \\
BLIP-large \cite{li2022blip}      & 384$\times$384                    & \multicolumn{1}{c|}{-}             & -                     & \multicolumn{1}{c|}{303M}   & \multicolumn{1}{c}{142M}                   \\
DFN \cite{fang2023dfn}           & 378$\times$378            & \multicolumn{1}{c|}{-}                                & \multicolumn{1}{c|}{-} & \multicolumn{1}{c|}{633M}       & \multicolumn{1}{c}{302M}         \\
SigLIP-So400m \cite{zhai2023siglip}  & 384$\times$384          & \multicolumn{1}{c|}{Web, OCR}                                & \multicolumn{1}{c|}{29B} & \multicolumn{1}{c|}{428M}       & \multicolumn{1}{c}{449M}       \\
InternVL-14B-224px \cite{chen2024internvl}   & 224$\times$224        & \multicolumn{1}{c|}{Doc, Chart, Natural}                                & \multicolumn{1}{c|}{1.4M} & \multicolumn{1}{c|}{5.9B}       & \multicolumn{1}{c}{7.8B}        \\ \cmidrule(r){1-1}\cmidrule(l){2-6}
Donut \cite{kim2022donut}        & 2560$\times$1920                  & \multicolumn{1}{c|}{Synthetic, Doc}        & 13M                   & \multicolumn{1}{c|}{74M}    & \multicolumn{1}{c}{128M}                   \\
Nougat \cite{blecher2023nougat}  & 896$\times$672                    & \multicolumn{1}{c|}{Doc}                   & 8.2M                  & \multicolumn{1}{c|}{74M}    & \multicolumn{1}{c}{275M}                      \\
Pix2Struct \cite{lee2023pix2struct}  & $2^{19}$ (shape-variable)         & \multicolumn{1}{c|}{Web}               & 80M                   & \multicolumn{1}{c|}{91M}    & \multicolumn{1}{c}{190M}                     \\
Vary \cite{wei2023vary}       & 1024$\times$1024                  & \multicolumn{1}{c|}{Doc, Chart, Natural}      & 8M                    & \multicolumn{1}{c|}{0.4B}   & \multicolumn{1}{c}{7.7B}                   \\
DocOwl1.5 \cite{hu2024mplug1.5}  & 448$\times$448 ($\times$9 crops)  & \multicolumn{1}{c|}{Doc, Table, Chart, Web, Natural} & 4M          & \multicolumn{1}{c|}{0.3B}   & \multicolumn{1}{c}{7.7B}                      \\
UReader \cite{ye2023ureader}  & 224$\times$224 ($\times$20 crops) & \multicolumn{1}{c|}{Doc, Table, Chart, Web, Natural} & 0.1M           & \multicolumn{1}{c|}{0.3B}   & \multicolumn{1}{c}{6.7B}                    \\
TextMonkey \cite{liu2024textmonkey}  & 896$\times$896            & \multicolumn{1}{c|}{Doc, Table, Chart, Scene Text}   & 2.5M            & \multicolumn{1}{c|}{1.9B}   & \multicolumn{1}{c}{7.7B}               \\
Qwen2-VL \cite{wang2024qwen2vl}           & \multicolumn{1}{c|}{Dynamic}             & \multicolumn{1}{c|}{Doc, Table, Chart, Web, OCR}                                & \multicolumn{1}{c|}{-} & \multicolumn{1}{c|}{0.7B}       & \multicolumn{1}{c}{7.6B}        \\ 
DSE \cite{ma2024unifying}  & \multicolumn{1}{c|}{1344$\times$1344}             & \multicolumn{1}{c|}{Web}                                & \multicolumn{1}{c|}{1.3M(FT)} & \multicolumn{1}{c|}{0.4B}       & \multicolumn{1}{c}{3.6B}      \\
ColPali \cite{faysse2024colpali} & \multicolumn{1}{c|}{448$\times$448}          & \multicolumn{1}{c|}{Doc, Table, Chart, Web}                                & \multicolumn{1}{c|}{12.7K(FT)} & \multicolumn{1}{c|}{0.4B}       & \multicolumn{1}{c}{2.5B}      \\ \bottomrule
\end{tabular}%
}
\label{tab:baseline_models}
\end{table}
\vspace{-10px}

\vspace{-3pt}
\subsection{Evaluation Protocol}
\vspace{-3pt}

\paragraph{Metrics.} We use MRR@$k$ and Recall@$k$ metrics to assess the retrieval performance on the NL-DIR benchmark, where $k$ is set to 1 and 10 for evaluation.

\begin{itemize}[itemsep=0pt,topsep=2pt,leftmargin=20pt]
    \item \textbf{MRR} stands for Mean Reciprocal Rank, which is calculated by the reciprocal of the golden label’s ranking in candidates. 
    \item \textbf{Recall} assesses the accuracy of the retrieval system by checking whether the golden label is present within the Top-k ranked results.
\end{itemize}

\begin{figure}[!htb]
\centering
\includegraphics[width=0.95\linewidth]{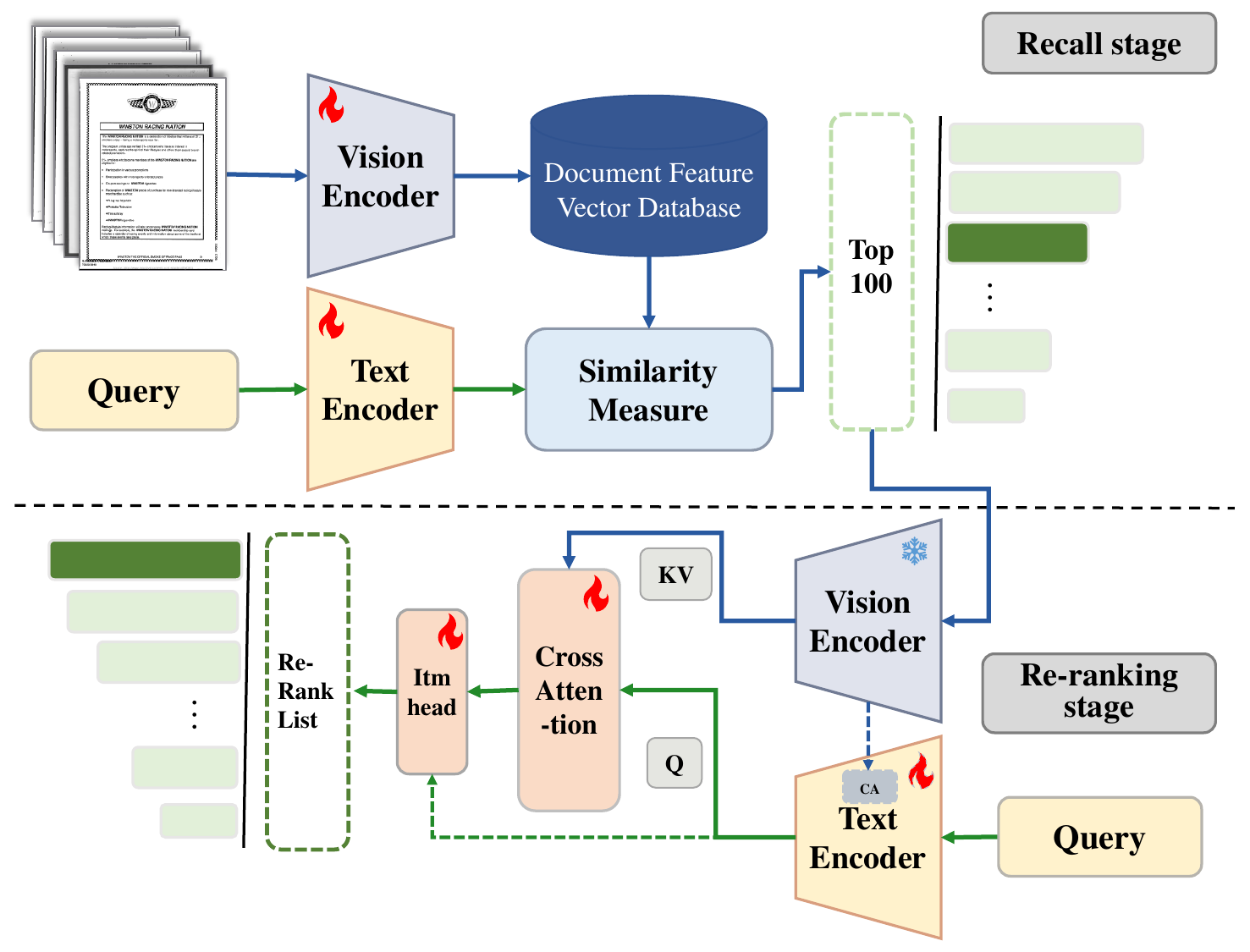}
\vspace{-10px}
\caption{The proposed two-stage approach includes the recall stage and re-ranking stage. The dashed borders and arrows represent the structure and flow of different models in the experiment.}
\label{fig:model}
\vspace{-20px}
\end{figure}

In this paper, following common retrieval paradigms, we adopt a two-stage approach: the recall stage is intended to achieve the rapid matching of large batches of document images; the re-ranking stage \cite{tan2021instance,lee2022correlation} is designed for precise reordering of candidate images. As shown in \cref{fig:model}, ``CA'' represents the cross attention module, while ``Itm head'' refers to a binary matching classification head.

\textbf{Recall stage.} In the recall stage, we conduct zero-shot and fine-tuning evaluations on the baseline models. The ranking score is calculated as the dot product between the visual and text representations.

During zero-shot evaluation, for contrastive VLMs, we directly extract the representation with the textual and visual encoders. For generative VDU models, visual representation is obtained from the visual module, while text representations are derived from the last decoder layer. We further impose both mean pooling to get the final representations. Notably, following two concurrent works, we maintain their original settings by using the End-of-Sequence (EOS) token of the LVLMs or compressing the feature dimensions of the last layer to generate the visual and textual representations.

During fine-tuning evaluation, we use batched query-image pairs from the training set to align the text and visual encoders through contrastive learning. The primary objective is to evaluate the representation capacity of visual encoders in various generative VDU models. However, the original text encoders in these models often exhibit limitations in both efficiency and alignment capability. To address this, we apply LoRA \cite{hu2021lora} and an extra alignment layer to the CLIP and BLIP text encoders, adapting them to align with the frozen visual encoders of different VDU models. We employ mean pooling for visual representations, while text encoders retain their original single-vector representations. For CLIP and BLIP, we optimize using InfoNCE loss. For SigLIP, which demonstrates strong zero-shot performance, we fine-tune both its text and visual encoders using LoRA and Sigmoid loss. 




\textbf{Re-ranking stage.} After the recall stage, we obtain the top 100 document images for each query as the initial ranking result, generated by SigLIP. In the re-ranking stage, we refine this ranking by re-ranking the top 100 retrieved results. A robust re-ranking model necessitates fine-grained interactions between image and text features. To achieve this, we fine-tune the cross-attention modules in models such as BLIP-ITM and Pix2Struct or incorporate additional cross-attention modules. To improve the learning process, we implement a hard negative mining strategy to identify challenging negative document images corresponding to the queries from the recall stage. Ultimately, we discover that incorporating additional cross-attention modules after the encoders, along with a combination of pointwise and pairwise loss, specifically targeting the top 10 hard negatives obtained during the recall stage for training, can significantly enhance performance during the re-ranking stage.

\vspace{-5pt}
\section{Result Analysis}

Following the above settings and evaluation protocol, we perform the two-stage approach on the test set of NL-DIR. The performance of different models is discussed here.

\vspace{-3pt}
\subsection{Analysis on Recall Stage}
\vspace{-3pt}

We provide a detailed analysis of the baseline models' recall performance under the zero-shot and fine-tuning settings.

\vspace{-10px}
\paragraph{Zero-shot Setting.} In \cref{table:recallzero-shot}, we evaluate the zero-shot performance of current contrastive VLMs and generative VDU models. As contrastive VLMs focus on the image-text matching task during pre-training, they have the potential to be competent for the real-world DIR task. Notably, SigLIP-So400m achieves the best zero-shot performance, with a recall@10 score of 61.18, significantly surpassing other models, attributed to its pre-train of a vast corpus of image-OCR pairs. Comparing all models in \cref{table:recallzero-shot}, it is evident that as the visual encoder expands, image input resolution increases, and additional training on image-text retrieval datasets occurs, the zero-shot retrieval performance of contrastive VLMs gradually improves. 

In contrast, the VDU models demonstrate inferior zero-shot retrieval performance. This is because the aforementioned VDU models are primarily designed for generative tasks (e.g., key information extraction and visual question answering) without specifically aligning representations. Nevertheless, generative models such as InternVL-14B-224px, after aligning the visual and text representations during pre-training, attain a noteworthy recall@10 score of 43.45. This score is considerably superior to the strongest generative VDU model, Qwen2-VL. 


\begin{table}[!htb]
\caption{The zero-shot results of contrastive VLMs and generative VDU models on the NL-DIR benchmark.}
\vspace{-10px}
\centering
\resizebox{0.48\textwidth}{!}{
\begin{tabular}{l|l|c|c|c}
\toprule
                                   & Models             & Recall@1 & Recall@10 & MRR@10 \\ \midrule
\multirow{7}{*}{\parbox{1.5cm}{Contrastive \\ VLMs}} 
                                   & CLIP-base          & 1.44     & 3.99      & 2.11                        \\ 
                                   & BLIP-base          & 2.54     & 6.02      & 3.48                        \\ \cmidrule{2-5} 
                                   & BLIP-large-384     & 3.84     & 10.68     & 5.66                        \\ 
                                   & BLIP-large-COCO    & 5.95     & 13.80     & 8.14                        \\ \cmidrule{2-5}
                                   & InternVL-14B-224px & 24.25    & 43.45     & 29.92                       \\
                                   & DFN                & 28.48    & 51.24     & 35.36                       \\
                                   & SigLIP-So400m      & \textbf{36.17} & \textbf{61.18} & \textbf{43.78}    \\ \midrule
\multirow{8}{*}{\parbox{1.5cm}{Generative VDU}}    
                                   & Donut              & 0.02     & 0.21      & 0.07                        \\
                                   & Nougat             & 0.01     & 0.23      & 0.07                        \\
                                   & Pix2Struct         & 0.02     & 0.21      & 0.07                        \\ \cmidrule{2-5} 
                                   & Vary               & 0.01     & 0.27      & 0.06                        \\
                                   & TextMonkey         & 0.02     & 0.22      & 0.07                        \\
                                   & DocOwl1.5          & 0.10     & 0.94      & 0.29                        \\
                                   & UReader            & 0.18     & 1.19      & 0.41                        \\
                                   & Qwen2-VL            &\textbf{0.29}  & \textbf{1.66}   & \textbf{0.59}    \\
\bottomrule
\end{tabular}
}
\vspace{-10px}
\label{table:recallzero-shot}
\end{table}

\vspace{-10px}
\paragraph{Fine-tuning Setting.} \cref{table:finetunealgnment} presents the results of aligning CLIP and BLIP with VDU models. The results indicate that models' performance is often positively correlated with the size of the visual encoder and pre-trained data. Notably, despite the limitations in the capacities of visual and text encoders, Pix2Struct has pre-trained on the largest dataset and has variable resolution document image input, so it still demonstrates satisfactory retrieval performance. Furthermore, we observe that BLIP's text representation is more suitable for the DIR task compared to CLIP, which reveals that the retrieval ability learned on natural image-text datasets can also be transferred to the document domain.

\begin{table}[!htb]
\caption{The retrieval results after aligning the visual encoders of the VDU models with the text encoders of CLIP and BLIP.}
\centering
\vspace{-10px}
\resizebox{0.43\textwidth}{!}{
\begin{tabular}{l|l|c|c|c}
\toprule
Text                    & Visual          & Recall@1       & Recall@10      & MRR@10         \\ \midrule
\multirow{8}{*}{CLIP}   & Donut           & 1.46           & 9.83           & 3.42           \\
                        & Nougat          & 1.65           & 9.50           & 3.48           \\
                        & Pix2Struct      & 4.15           & 20.01          & 8.18           \\ \cmidrule{2-5} 
                        & Vary            & 2.11           & 13.23          & 4.76           \\
                        & DocOwl1.5       & 4.19           & 20.18          & 8.22           \\
                        & UReader         & 4.40           & 21.09          & 8.62           \\
                        & TextMonkey      & \textbf{4.46}  & \textbf{21.99} & \textbf{8.92}  \\ \midrule
\multirow{8}{*}{BLIP}   & Donut           & 1.65           & 10.30          & 3.68           \\
                        & Nougat          & 1.57           & 10.22          & 3.60           \\
                        & Pix2Struct      & \textbf{5.07}  & 21.85          & 9.34           \\ \cmidrule{2-5} 
                        & Vary            & 2.49           & 14.03          & 5.23           \\
                        & DocOwl1.5       & 4.15           & 20.13          & 8.18           \\
                        & UReader         & 4.54           & 22.21          & 9.00           \\
                        & TextMonkey      & 5.03           & \textbf{23.33} & \textbf{9.60}  \\ 
\bottomrule
\end{tabular}
}
\vspace{-20px}
\label{table:finetunealgnment}
\end{table}

Because of SigLIP's outstanding performance in the zero-shot setting, we perform an independent evaluation of its capabilities. As shown in \cref{table:siglipalign}, we attempt to fine-tune its text module and observe an increase in recall@10 from 61.18 to 79.40, demonstrating strong transferable performance on NL-DIR. Additionally, aligning the textual encoder with the powerful visual encoder of Pix2Struct gives results that exceed results in \cref{table:finetunealgnment}, but is inferior to the original SigLIP model. This highlights the importance of aligning visual and text representations in the common semantic space. Finally, we fine-tune its visual module (i.e., SigLIP-Image-LoRA), resulting in the best performance during the recall stage. It's worth noting that we also test the best model's recall@100 score, which reaches 97.52, indicating a strong ability to recall document images. This will facilitate the subsequent re-ranking stage.

\vspace{-10px}
\begin{table}[!htb]
\caption{The retrieval results of SigLIP after fine-tuning.}
\centering
\vspace{-10px}
\resizebox{0.45\textwidth}{!}{
\begin{tabular}{c|l|c|c|c}
\toprule
\multicolumn{1}{l|}{Text} & Visual            & Recall@1       & Recall@10      & MRR@10         \\ \midrule
\multirow{3}{*}{SigLIP-Text-LoRA}   & SigLIP            & 54.32          & 79.40          & 62.39          \\
                          & Pix2Struct        & 13.14          & 41.57          & 20.98          \\
                          & SigLIP-Image-LoRA & \textbf{69.33} & \textbf{89.72} & \textbf{76.29} \\ 
\bottomrule
\end{tabular}
}
\vspace{-10px}
\label{table:siglipalign}
\end{table}

\vspace{-3pt}
\subsection{Analysis on Re-Ranking Stage} 
\vspace{-3pt}

We utilize the cross attention for fine-grained interaction between images and text, allowing us to re-rank the top 100 results during the recall stage. In \cref{table:reranker}, we explore the combination of various visual and text encoders, as well as different interaction methods, and compare our approach with existing works based on text retrieval and LVLMs. Due to the image-text matching (ITM) pre-training tasks, BLIP-ITM demonstrates better performance in the re-ranking stage compared to SigLIP. However, because of the limitations of the visual encoder's representational capacity and the gap between the training data and pre-training data, the re-ranking process results in even negative optimization compared to the original recall results. To address this, we leverage Pix2Struct to obtain the fine-grained representation of images. Regardless of whether we add extra cross-attention or utilize the original cross-attention of Pix2Struct for the ITM task, all approaches yield significant improvements in re-ranking results. 

\vspace{-10pt}
\begin{table}[!htb]
\caption{The performance of settings with different visual encoders, text encoders, and cross-attention during the re-ranking stage, as well as the experimental results of existing text-based and LVLM retrieval methods.``ZS'' and ``FT'' are short for zero-shot and fine-tuning respectively.}
\centering
\vspace{-10px}
\resizebox{0.48\textwidth}{!}{
\begin{tabular}{c|l|l|l|c|c|c}
\toprule
\multicolumn{1}{l|}{} & Text       & Visual     & Cross attention & Recall@1       & Recall@10      & MRR@10         \\  \midrule
\multirow{3}{*}{ZS}   
                      & OCR-IR     & -          & -               & 52.83          & 71.63          & 58.85          \\
                      & DSE        & DSE        & -               & 69.43          & 87.57          & 75.61          \\
                      & ColPali    & ColPali    & -               & \underline{79.65}    & \underline{91.64}    & \underline{83.79}    \\ \midrule 
\multirow{5}{*}{FT}   & SigLIP     & SigLIP     & Extra     & 1.20           & 12.98          & 3.73           \\
                      & BLIP-ITM   & BLIP-ITM   & Original             & 13.85          & 27.24          & 17.15          \\
                      & SigLIP     & Pix2Struct & Extra     & 25.99          & 79.40          & 42.35          \\
                      & Pix2Struct & Pix2Struct & Original             & 72.68          & 84.63          & 76.68          \\
                      & BLIP-ITM   & Pix2Struct & Extra     & \textbf{81.03} & \textbf{94.17} & \textbf{85.68} \\ \bottomrule
\end{tabular}
}
\vspace{-10px}
\label{table:reranker}
\end{table}

For the two concurrent works, since they are already fine-tuned on their respective document retrieval datasets, we employ zero-shot evaluation for comparison. Compared to DSE \cite{ma2024unifying}, although ColPali \cite{faysse2024colpali} has less training data, it achieves impressive results on retrieval tasks through fine-grained interactions within the same representation space. After achieving fine-grained alignment using the smaller BLIP text representation alongside the Pix2Struct visual representation, we attain a recall@1 of 81.03, producing results comparable to those of the LVLMs-based method. While existing re-ranking and LVLM-based methods perform well on this dataset, the proposed dataset remains valuable for zero-shot evaluation of foundation models and benchmarking other tasks after training.



We also report the results of the OCR-IR pipeline here for reference. In detail, this pipeline uses the commonly used Tesseract OCR\footnote{\url{https://github.com/tesseract-ocr/tesseract}} to extract text from document images, concatenates text content with spaces, and adopts the currently competitive text retrieval model BGE \cite{xiao2023bge} for text document retrieval. Based on Pixel2Struct, the enhanced version outperforms OCR-IR, showing the potential of developing OCR-free models to address this task.


\textbf{Bad cases.} We attempt to fine-tune DSE on the training dataset. However, the transfer results are not satisfactory, probably due to the size and diversity of the fine-tuning data and the limitation of batch size. Additionally, regarding the visual encoder, we explore using more comprehensive visual information for fine-grained re-ranking, such as TextMonkey and Pix2Struct-Large, but do not achieve better results, while the encoding time is significantly increased. We have also tried to use LoRA to fine-tune the text encoder during the re-ranking stage alongside cross-attention, while the results were not promising. For the additional layers of cross-attention, our optimal configuration is one layer, as increasing the number of layers makes model convergence more challenging. In terms of model loss, removing the pairwise loss leads to a slight decrease in performance, while eliminating the pointwise loss results in much slower convergence and a more pronounced drop in performance.

\vspace{-3pt}
\subsection{Evaluating Different Query and Image Subsets}
\vspace{-3pt}

\paragraph{Query Type Subsets.} To demonstrate the impact of different types of queries, we sample from the test set based on the length of queries and the number of overlapping words with the original OCR annotations from the source data. As a result, queries with a length greater than 18 and more than 10 overlapping words with the original OCR are categorized as a concrete set, which contains 3100 distinct queries. These queries typically carry more precise semantic information about the document. Queries with a length shorter than 14 and fewer than 6 overlapping words with the original OCR were categorized as an abstract set, also containing 3,100 queries. The queries in the subset are generally more brief and abstract. The performance of the two different query types on the test set is shown in \cref{table:query_results}.

\vspace{-10px}
\begin{table}[!htb]
\caption{Retrieval results of different query categories on the NL-DIR benchmark.}
\centering
\vspace{-10px}
\resizebox{0.48\textwidth}{!}{%
\begin{tabular}{l|l|c|c|c}
\toprule
\multicolumn{1}{l|}{Query Type} & Stage & Recall@1& Recall@10& MRR@10 
\\ \midrule
\multirow{2}{*}{Abstract}   & Recall    & 65.37  & 87.76      
& 72.99  
\\ 
                            & Re-ranking    & 68.18  & 
90.56      
& 75.73  
\\ \midrule
\multirow{2}{*}{Concrete}   & Recall    & 86.23  & 97.22      
& 90.12  
\\ 
                            & Re-ranking    & \textbf{95.10}  & 
\textbf{99.35}      
& \textbf{96.66}  
\\ \midrule
\multirow{2}{*}{Org test}   & Recall    & 69.33  & 89.72      
& 76.29  
\\ 
                            & Re-ranking    & 81.03  & 
94.17      & 85.68  \\
\bottomrule
\end{tabular}%
}
\vspace{-10px}
\label{table:query_results}
\end{table}

In terms of retrieval performance across both two stages, the concrete set outperforms the original test set, which in turn outperforms the abstract set. For example, in the MRR@10 metric, the concrete set achieves a score of 96.66, which is higher than the original test set at a score of 85.68, and the abstract set at a score of 75.73. Fine-grained re-ranking improves performance for all queries, with the most significant enhancements observed for query sets with more detailed intents.

\vspace{-10px}
\paragraph{Visually Rich Document Subsets.} Based on the original document labels from the Industry Documents Library website, we attempt to extract two subsets of visually rich document images from the test set: advertisement and chart. These subsets contain 61 and 163 images, respectively, along with 305 and 815 queries. We conduct experiments on these two subsets, with the results shown in \cref{table:performance_metrics}.

\vspace{-5px}
\begin{table}[!htb]
\caption{Retrieval performance of the model on two visually rich document evaluation subsets.}
\vspace{-10px}
\centering
\resizebox{0.48\textwidth}{!}{%
\begin{tabular}{l|l|c|c|c}
\toprule
\multicolumn{1}{l|}{Category}     & Method & Recall@1& Recall@10& MRR@10 
\\ \midrule
\multirow{3}{*}{Chart}        & Recall    & 89.20  & 99.02& 93.13  
\\ 
                              & Re-ranking    & 89.70  & 98.65& 92.91  
\\ 
                              & OCR-IR & 74.96  & 92.39& 80.57  
\\ \midrule
\multirow{3}{*}{Advertisement} & Recall   & 87.64  & 99.01& 92.17  
\\ 
                               & Re-ranking    & 80.66  & 97.05& 85.55  
\\ 
                               & OCR-IR & 66.88  & 85.90& 72.40  \\ 
\bottomrule
\end{tabular}%
}
\vspace{-10px}
\label{table:performance_metrics}
\end{table}

From the results, the Recall@10 performance of the chart subset and the advertisement subset can achieve about 99 at the recall stage, suggesting that the model has a basic retrieval capability for visually rich document images after image-text contrast fine-tuning. During the re-ranking stage, we conduct an in-depth interaction with fine-grained text information and document images. While this approach somewhat weakens the understanding and matching capabilities for certain types of document images, such as those resembling advertisements. Nevertheless, it still outperforms the OCR-IR pipeline overall on these two subsets.

\vspace{-3pt}
\subsection{Time and Storage Analysis}
\vspace{-3pt}

In contrast to the text-based or LVLM-based method, the two-stage retrieval approach proposed in this paper often demonstrates superior performance in terms of offline encoding, storage capacity, and online query performance, while also presenting more space for optimization.

\textbf{Offline Encoding.} For the evaluation of offline encoding speed, we utilize 100 sampled images for offline feature extraction, employing a 4090 GPU with a batch size of 4. The average encoding time per image for SigLIP is 0.07 seconds. For text-based methods, the OCR module typically contributes significantly to the overall time consumption, leading to an average encoding time of 2.56 seconds. For LVLM-based methods, the encoding times for DSE and ColPali are 0.62 seconds and 0.65 seconds, respectively. Due to the constraints imposed by model size, their offline encoding times are relatively long. Therefore, a well-performing contrastive VLM in the recall stage can effectively reduce encoding time.

\textbf{Storage Capacity.} In practical scenarios, the storage space required for retrieved documents is a significant consideration. The embedding size for SigLIP is 4KB, which is comparable to the storage requirements of BGE at 3KB and DSE at 6KB. In contrast, ColPali requires more storage space at 256KB due to its use of multi-vector embeddings. In the recall stage, using single-vector dense representations can save a significant amount of storage space.

\textbf{Online Query Performance.} In the course of our experiments, we observe that the online encoding time for queries using LVLM is relatively prolonged. The online retrieval efficiency is similar across different models. During the re-ranking stage following recall in our model, the time to generate visual embeddings is 0.2 seconds per image. However, we can optimize the processing of the top 100 results by utilizing parallel visual encoding, which will help reduce the overall time consumption. 

\vspace{-3pt}
\subsection{Qualitative Analysis}
\vspace{-3pt}

\begin{figure}[!htb]
\centering
\includegraphics[width=0.90\linewidth]{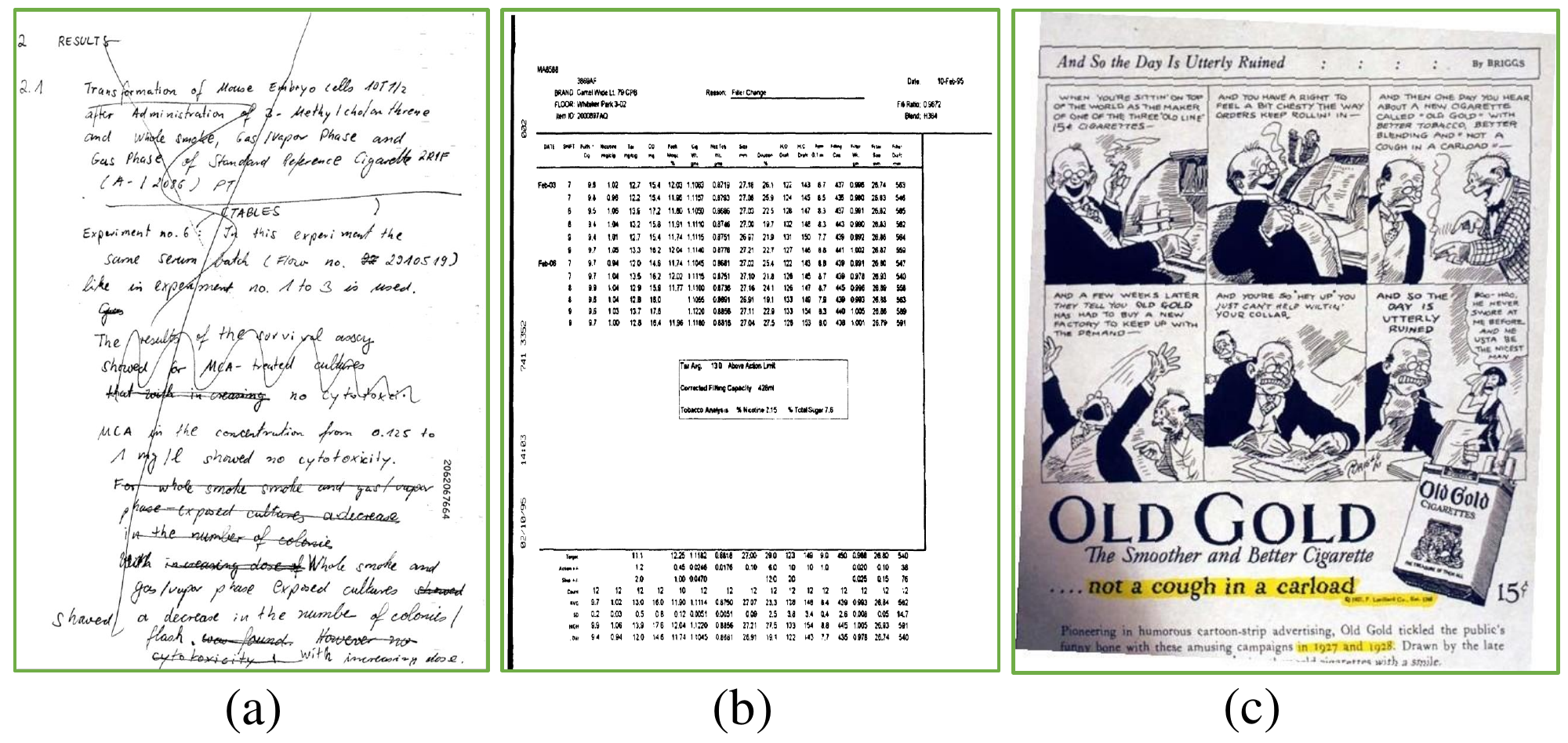}
\vspace{-10px}
\caption{OCR-free models perform exceptionally well in retrieving the following types of images: (a) Handwritten text. (b) Out-of-vocabulary. (c) Cartoon characters.}
\label{fig:goodcases}
\vspace{-10px}
\end{figure}

As shown in \cref{fig:goodcases}, compared to the OCR-dependent method, the OCR-free NL-DIR model exhibits superior performance in scenarios characterized by poor image quality (e.g., handwritten text), out-of-vocabulary problems (e.g., the query content is ``Whitalter'' but the OCR result is ``Whéaaker''), and visual information from non-text elements (e.g., the query mentions ``particularly its humorous cartoon-strip ads''). The model demonstrates the robust ability to understand visual information within documents, eliminating the negative impact of OCR errors.

\vspace{-5px}

\vspace{-5px}
\section{Conclusion}
\vspace{-5px}

In this paper, we introduce the NL-DIR task and present a corresponding dataset derived from real document scenarios. We employ a two-stage approach to build a new benchmark by evaluating the performance of existing contrastive VLMs and generative VDU models. The experiments are performed under both the zero-shot and fine-tuning settings, demonstrating the strengths and weaknesses of existing methods on NL-DIR. Substantial analysis and insights are provided to show the value of the proposed benchmark. We hope this work could facilitate the development of future research in the VDU area.

\vspace{-5px}
\section*{Acknowledgement}
\vspace{-5px}
This work is supported in part by National Key R\&D Program of China (No.2022YFB3103800) and National Natural Science Foundation of China (No.U2336205). Also supported by the fund of the Laboratory for Advanced Computing and Intelligence Engineering and the Collaborative Education Project of the Ministry of Education: Construction of Cyberspace Security Experimental Teaching and Training Platform (2408131129).
{
    \small
    \bibliographystyle{ieeenat_fullname}
    \bibliography{main}
}



\clearpage
\maketitlesupplementary

In the supplementary materials, we provide more details on dataset construction, along with the experimental training and testing processes. Additionally, we discuss further experimental findings and future work, concluding with an overview of the Licensing, Hosting, and Maintenance Plan, as well as the Datasheet.

\section{Dataset Details}

\subsection{Query Analysis}

\begin{figure}[!htb]
\begin{subfigure}{0.45\linewidth}
    \centering
    \includegraphics[width=\linewidth]{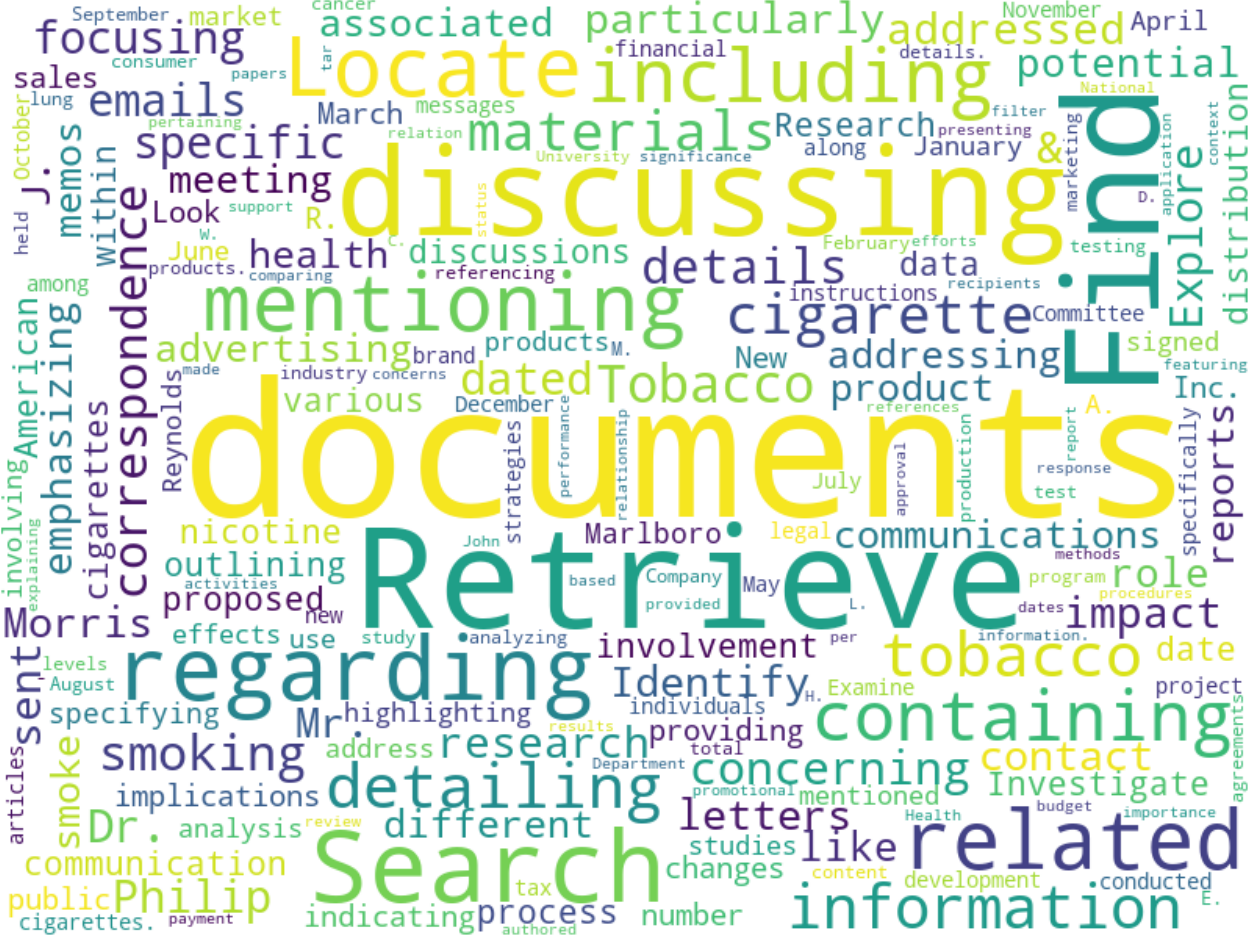}
    \caption{}
    \label{subfig:wordcloud}
\end{subfigure}
\hfill
\begin{subfigure}{0.45\linewidth}
    \centering
    \includegraphics[width=\linewidth]{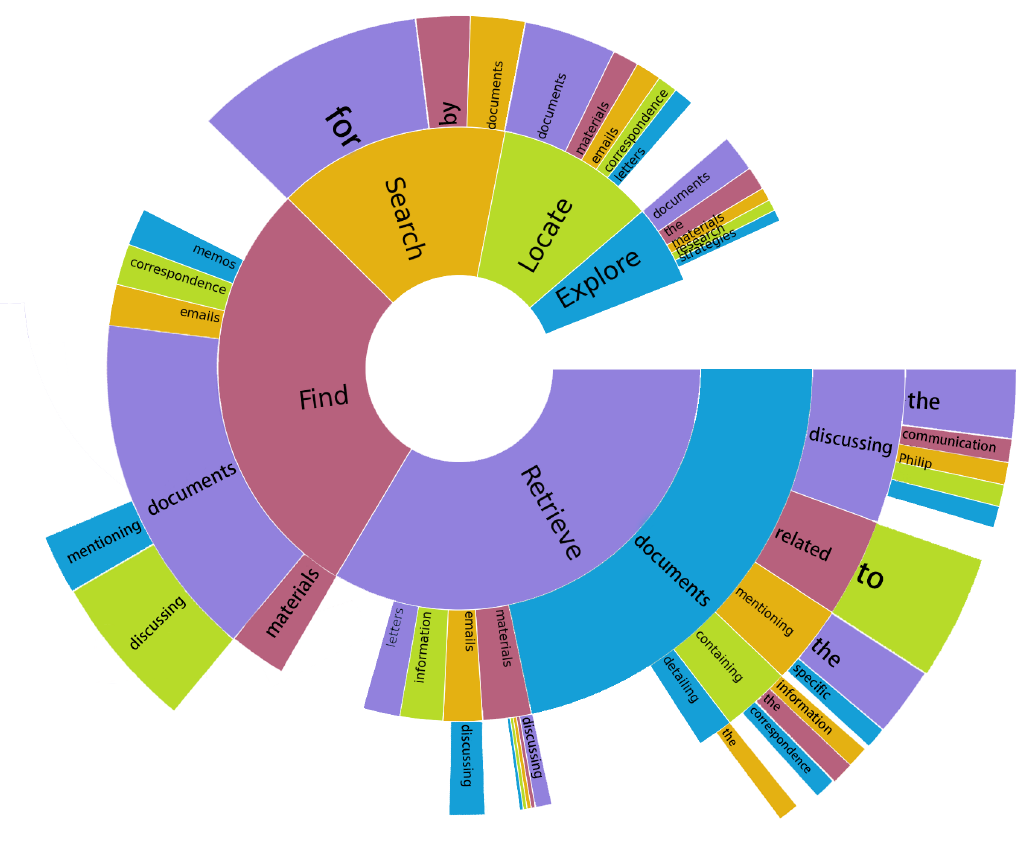}
    \caption{}
    \label{subfig:fourthwords}
\end{subfigure}
\vspace{-5px}
\caption{Query Analysis. (a) Word cloud of queries after removing stop words. (b) Distribution of first four words in queries in the NL-DIR dataset.}
\vspace{-10px}
\label{fig:wordcloud_fourthwords}
\end{figure}

We construct a word cloud in \cref{subfig:wordcloud} that removes stop words. The word cloud displays the most frequently occurring words and reveals the primary intention and key topics of retrieval. \cref{subfig:fourthwords} presents a sunburst diagram depicting the distribution of the first four words in the queries. The diagram reveals that queries frequently begin with words such as ``retrieve'', ``find'', and ``search'', which indicates the intention of information retrieval. The outer ring of the sunburst provides even more granular details, likely representing specific topics or types of information being sought, such as ``documents'', ``emails'', and ``memos''.

\begin{figure}[t]
\centering
\includegraphics[width=0.6\linewidth]{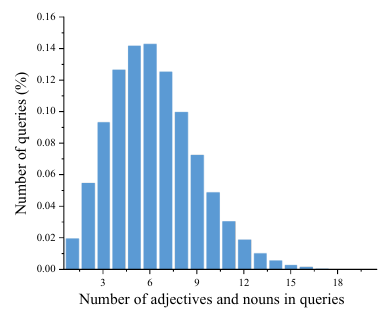}
\vspace{-5px}
\caption{Query length distribution after NER processing.}
\vspace{-10px}
\label{fig:difmethod}
\end{figure}

To extract more descriptive words, we process the queries using spacy's NER module to extract adjectives and nouns. The query length distribution, shown in Fig.\ref{fig:difmethod}, is concentrated between 3 and 9 words.

\subsection{Collection Details of the NL-DIR Dataset}
Initially, we attempt to build the Natural Language-Based Document Image Retrieval (NL-DIR) dataset by utilizing existing information retrieval datasets by rendering the documents into images. However, these images often fail to reflect the distribution of real-world documents. Therefore, we decide to construct the dataset by generating the corresponding queries.

We conduct extensive research and find datasets with a large scale of document images and relatively good OCR results. ``Relatively good OCR results'' refers to commercial OCR systems compared to the open-source Tesseract OCR. Finally, some images from DocVQA \cite{mathew2021docvqa} and OCR-IDL \cite{ocridl2022} are sampled to build the dataset. This allows us to obtain real-world document images with relatively good layout and content information as mentioned in the main paper. For OCR-IDL \cite{ocridl2022}, we collect and use the first page of its PDF files as a default choice, which contains richer semantic content. The subsequent construction process can be found in the main paper. In the following part, we also provide the prompts used in the query generation and filtering processes, as well as the standards for manual verification.

\subsection{Scoring Models and Manual Verification}
The scoring models used to pre-score the ten generated queries include a large language model (ChatGPT \cite{john2023chatgpt}), a multimodal large vision-language model (Qwen-VL-Plus \cite{bai2023qwen}), and two contrastive models (CLIP \cite{radford2021clip} and BLIP \cite{li2022blip}). The large language model (LLM) enables more effective analysis of the query content when combined with the OCR text, while the large vision-language model (LVLM) incorporates some visual elements from the document images for scoring. The final two models, which have undergone extensive pre-training on cross-modal image-text alignment, provide a preliminary score based on the degree of similarity between the text query and the document image. We collect and assign the aforementioned scores to the corresponding queries for each image.

\begin{figure}[t]
\centering
\includegraphics[width=0.9\linewidth]{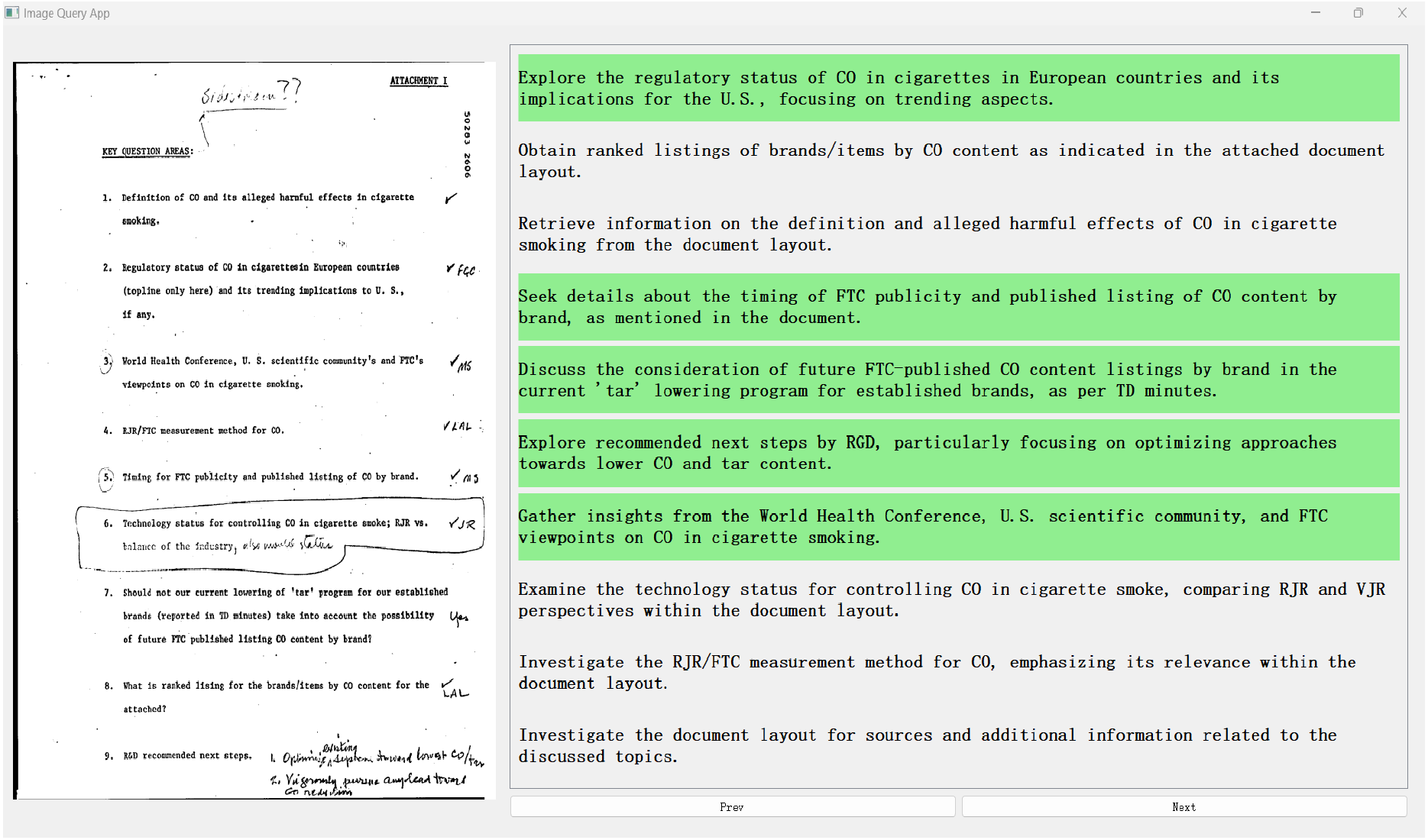}
\vspace{-5px}
\caption{Visualization for manual verification.}
\label{fig:vis_filiter}
\vspace{-10px}
\end{figure}

By designing and providing a visualization interface as \cref{fig:vis_filiter}, we display each image, the queries, and their scores for human verification. During the human verification process, we first remove damaged document images and inappropriate queries. The reserved pairs are filtered based on the above scores and query quality, ensuring that queries are as strongly related to the current image as possible. Then the annotators are asked to filter out ambiguous queries and images as much as possible.

Specifically, for document images, we will remove those with significant quality degradation or very low information content. For queries, we apply filtering rules to exclude those containing specific characters, such as queries that include ``UCSF'' or the original document source information. When both document images and queries are involved, we use the filtering methods mentioned in the body of the text to filter them accordingly.

To alleviate data bias, we strive to ensure consistent query quality across different document categories during manual filtering. Finally, through these two processes, we obtain a high-quality, fine-grained NL-DIR dataset.

\subsection{Visualized Examples of NL-DIR}
\begin{figure}[!htb]
\centering
\includegraphics[width=0.5\textwidth]{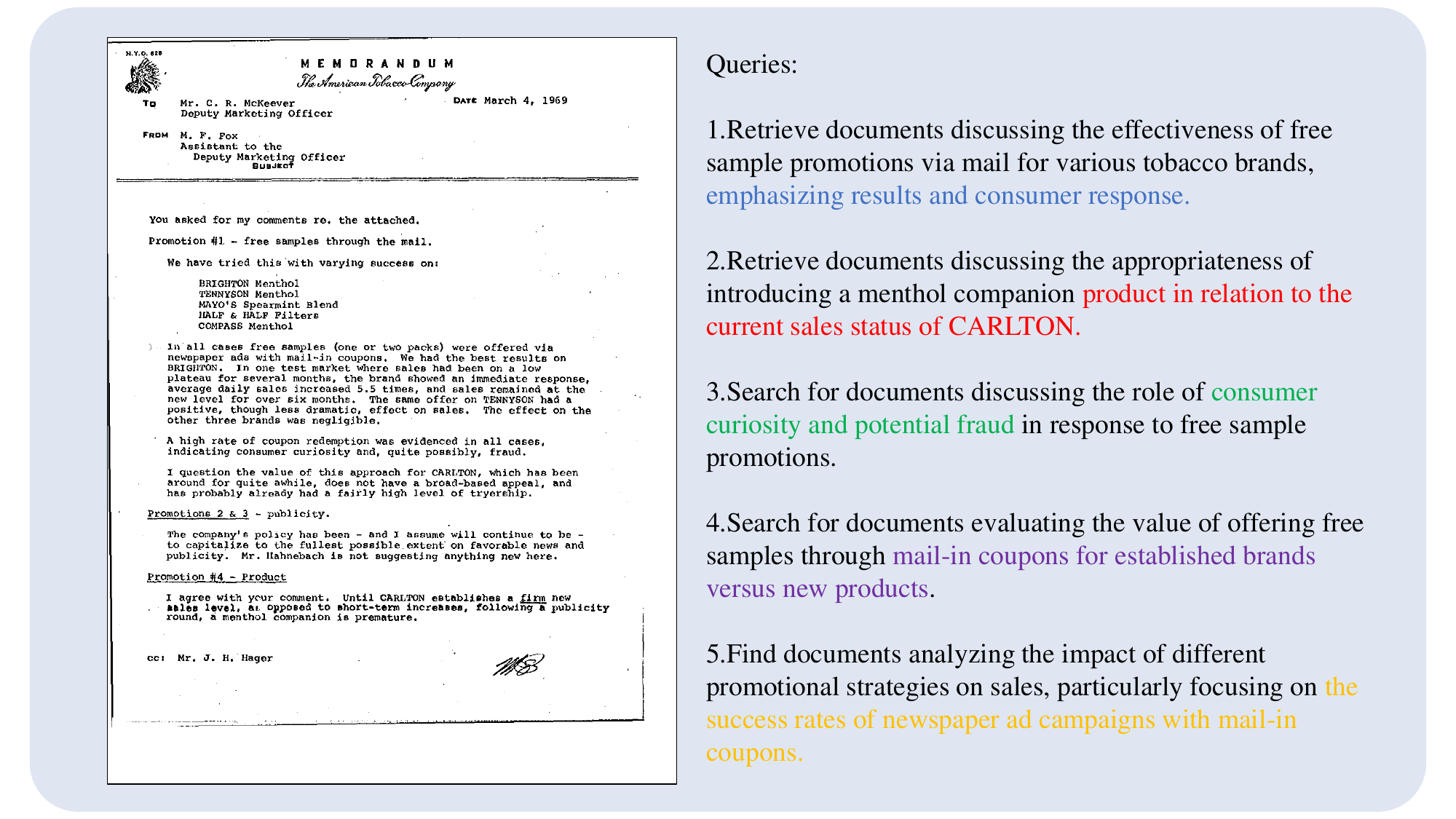}
\vspace{-10px}
\caption{Letter: when the textual information is abundant and there is relatively little structured information, the corresponding five queries mainly focus on understanding the content of the entire image.}
\label{fig:eg1}
\vspace{-10px}
\end{figure}

\begin{figure}[!htb]
\centering
\includegraphics[width=0.5\textwidth]{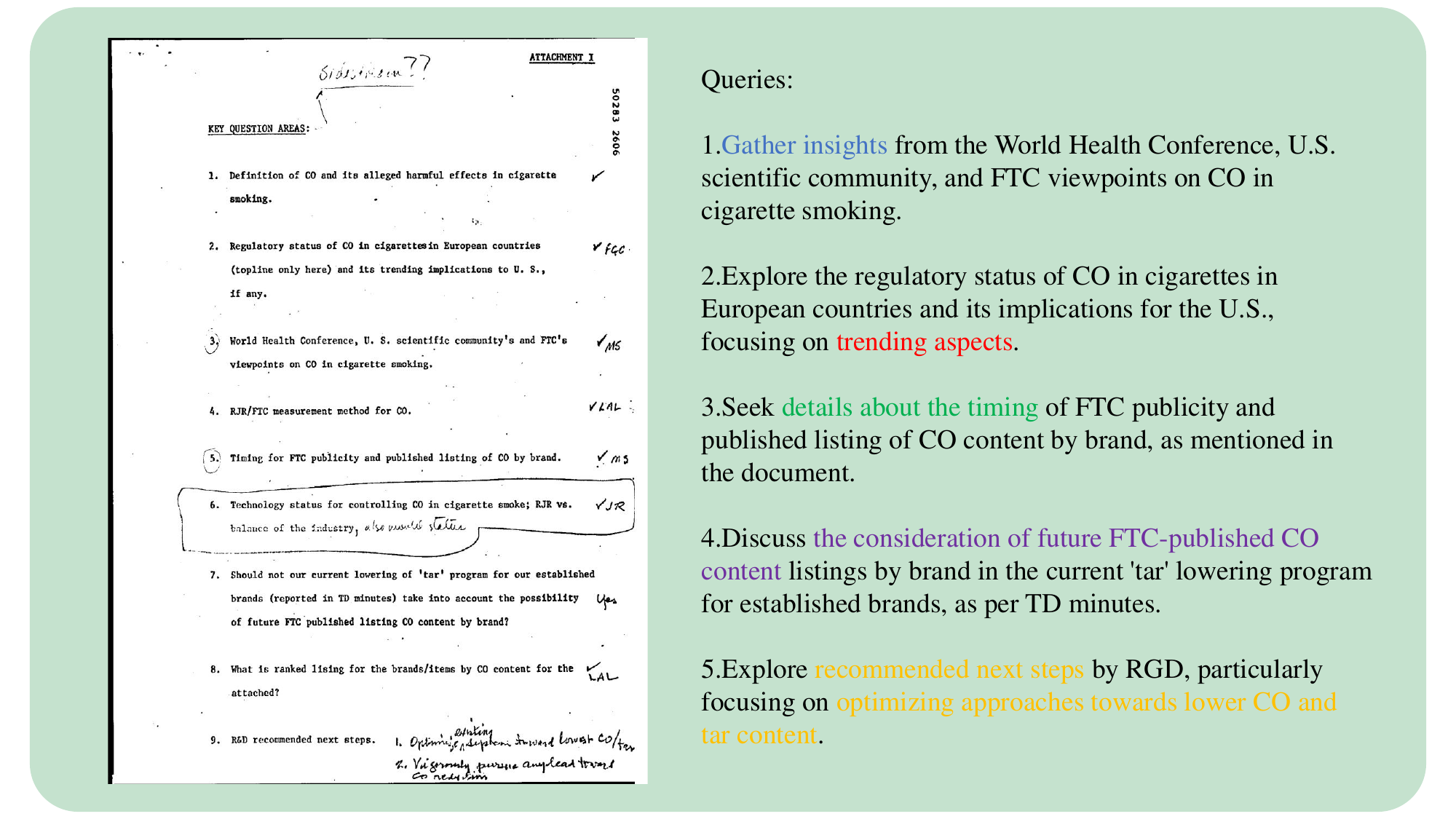}
\vspace{-10px}
\caption{Report: if the document image contains entries, the query is likely to be finely granular in its focus, with varying focus for each entry.} 
\label{fig:eg2}
\vspace{-10px}
\end{figure}

\begin{figure}[!htb]
\centering
\includegraphics[width=0.5\textwidth]{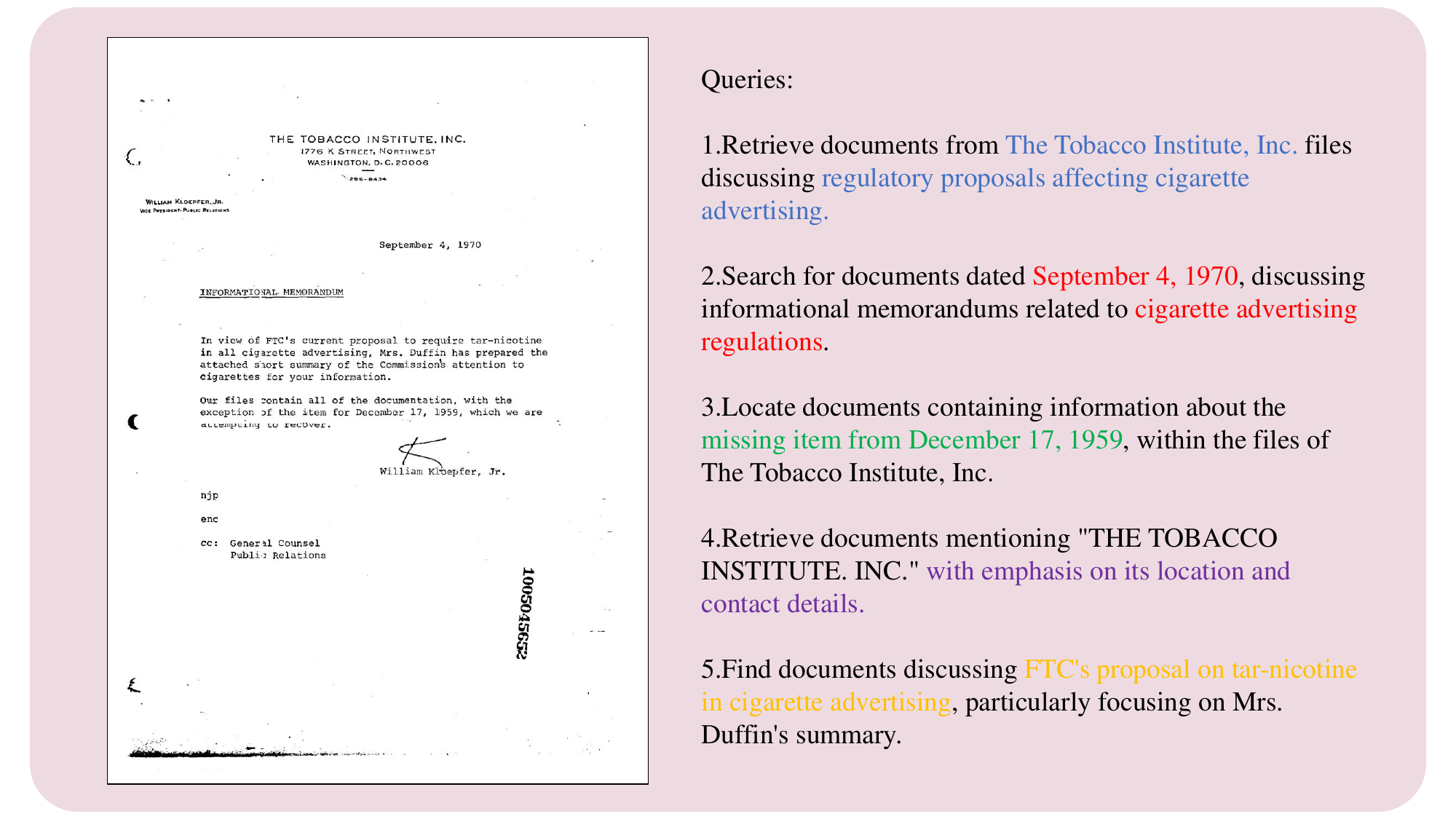}
\vspace{-10px}
\caption{Memo: it can be observed that the query will mine and search for some unique information within the document.} 
\label{fig:eg3}
\vspace{-10px}
\end{figure}

\begin{figure}[!htb]
\centering
\includegraphics[width=0.5\textwidth]{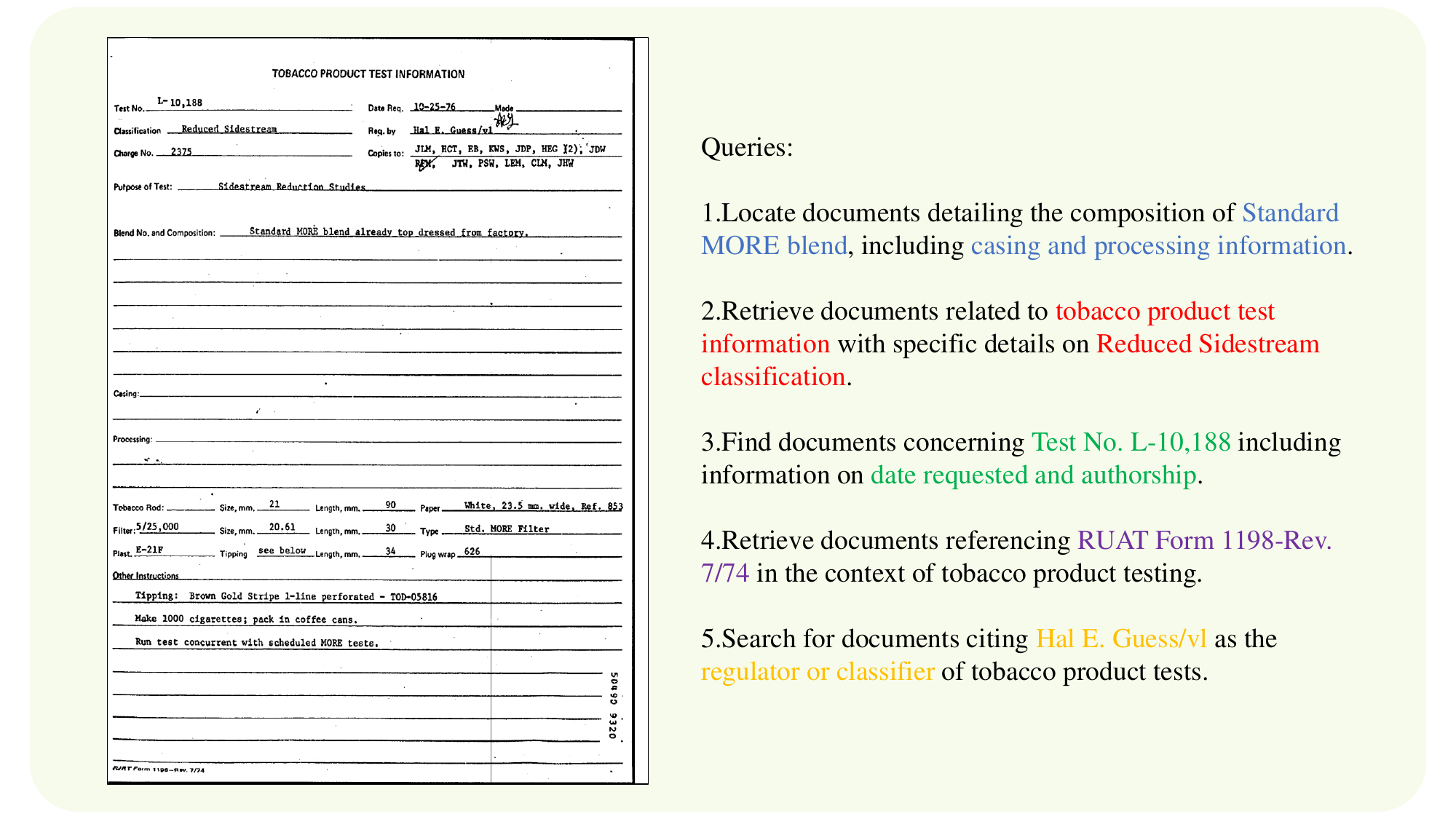}
\vspace{-10px}
\caption{Form: in the context of form images, greater emphasis is placed on the querying of the names and contents of different fields within the form.} 
\label{fig:eg4}
\vspace{-10px}
\end{figure}

\begin{figure}[!htb]
\centering
\includegraphics[width=0.5\textwidth]{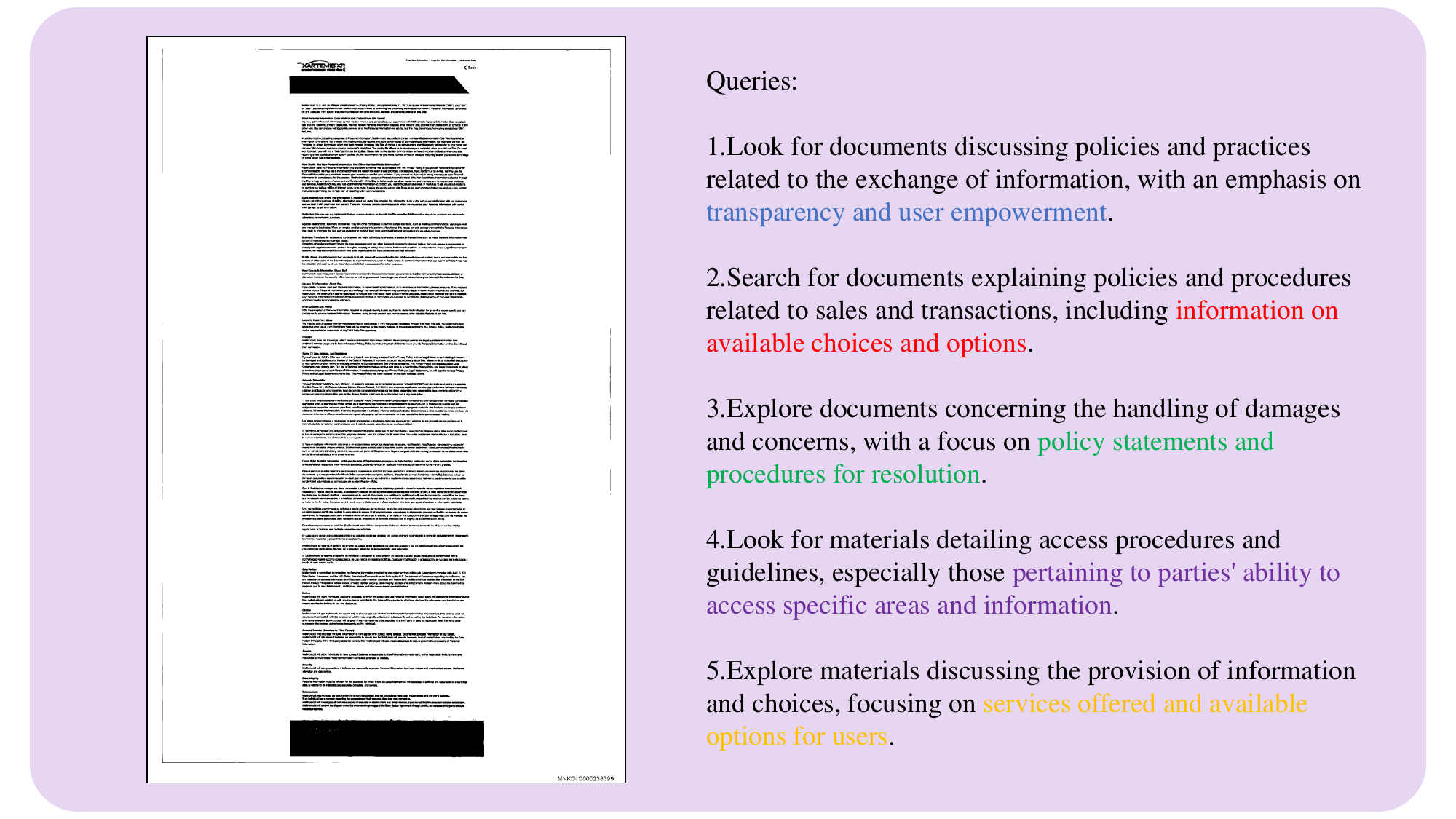}
\vspace{-10px}
\caption{Document: when the document image is rich in textual information, the query will be constructed to analyze the document as a whole and summarise its content.} 
\label{fig:eg5}
\vspace{-10px}
\end{figure}

This section presents examples and analysis of the five most common types of document images and their corresponding query statements. As shown in \cref{fig:eg1}, \cref{fig:eg2}, \cref{fig:eg3}, \cref{fig:eg4}, and \cref{fig:eg5}, the vocabulary related to document images is labeled with different colors in each query, showing the diversity of generated queries.

\section{Experimental Details}\label{sec:experiment_details}

\subsection{Recall and Re-Ranking Setting}

\paragraph{Recall stage.} In the zero-shot setting, for Contrastive VLMs, we directly extract their original visual representations. In contrast, for Generative VDU models, we take the final output from each visual encoder and perform mean pooling to obtain the visual representation. Both representations are stored in the FAISS vector library \footnote{\url{https://github.com/facebookresearch/faiss}}. During the extraction process for Generative VDU models, we also attempt to use the entire VLM to extract representations from the final output of the language layer as visual features. However, the retrieval performance is similar to that of directly extracting the visual encoder's representations, but the process take significantly longer. Therefore, we do not discuss this approach further. 
The dot product is then utilized to query the document image representation in the vector library, ultimately yielding zero-shot results.

In the fine-tuning setting, we use LoRA \cite{hu2021lora} to fine-tune the text encoders (i.e., CLIP \cite{radford2021clip} and BLIP \cite{li2022blip}), with parameters set to \( r=8 \) and \( \text{lora alpha}=16 \). After fine-tuning, we align these encoders with various VDU models. Linear layers are employed to fine-tune the mean pooling features, consisting of two layers with a residual connection that maps the original feature dimensions to 512. We also attempt to fine-tune both visual and text encoders simultaneously using LoRA; however, this approach not only significantly increase training time but also lead to a decrease in retrieval efficiency. 

For SigLIP \cite{zhai2023siglip}, we directly utilize its original model structure and apply LoRA to fine-tune both the text and visual encoders with parameters set as \( r=32 \), \( \text{lora alpha}=32 \), \( \text{weight decay}=1\times10^{-4} \), \( \text{warmup steps} = 2.5\% \), \( \text{lr}=5\times10^{-5} \). With a batch size of 32, fine-tuning for 10 epochs yields the best recall retrieval result.

\paragraph{Re-ranking stage.} We conduct a comparison with contemporaneous models, such as DSE \cite{ma2024unifying} and ColPali \cite{faysse2024colpali}, which utilize large visual-language models to encode queries and images. Considering factors like encoding time, storage space, retrieval efficiency, and training costs, we test these models in zero-shot setting. 

During the fine-tuning of the re-ranker, we primarily focus on the models that perform well in the recall stage. We use several models that have original cross-attention modules, specifically BLIP-ITM \cite{li2022blip} and Pix2Struct-base \cite{lee2023pix2struct}, or incorporate additional cross-attention for fine-tuning. For the pre-trained BLIP-ITM model, we fine-tune its language module directly. In the case of Pix2Struct-base, we add an additional ITM head and fine-tune the language module accordingly. For models with additional cross-attention, we enable interactions between the original fine-grained features to improve re-ranking results. The fine-tuning parameters for these models are set as follows: \( r = 32 \), \( \text{lora alpha} = 32 \), \( \text{lr} = 1 \times 10^{-3} \) (\( \text{lr} = 1 \times 10^{-4} \) for LoRA), \( \text{weight decay} = 1 \times 10^{-4} \). The optimizer follows a cosine decay schedule with \( T_{\text{max}} = 10 \) and \( \eta_{\text{min}} = 1 \times 10^{-5} \), and we use a batch size of 8. 

In the future, we will release the dataset with its construction code, evaluation code, model code, and weights to facilitate reproducibility for researchers.

\subsection{Case Analysis}

To better observe the results of fine-grained interactions during the re-ranking stage, we use the attention scores from the query and key in the cross-attention module to visualize the interactions between the query and the images. We aggregate the attention scores for each token in the query and superimpose the heatmap on top of the original image, allowing us to identify the regions in the image that are most relevant to the query. 

\begin{figure}[!htb]
\centering
\includegraphics[width=0.45\textwidth]{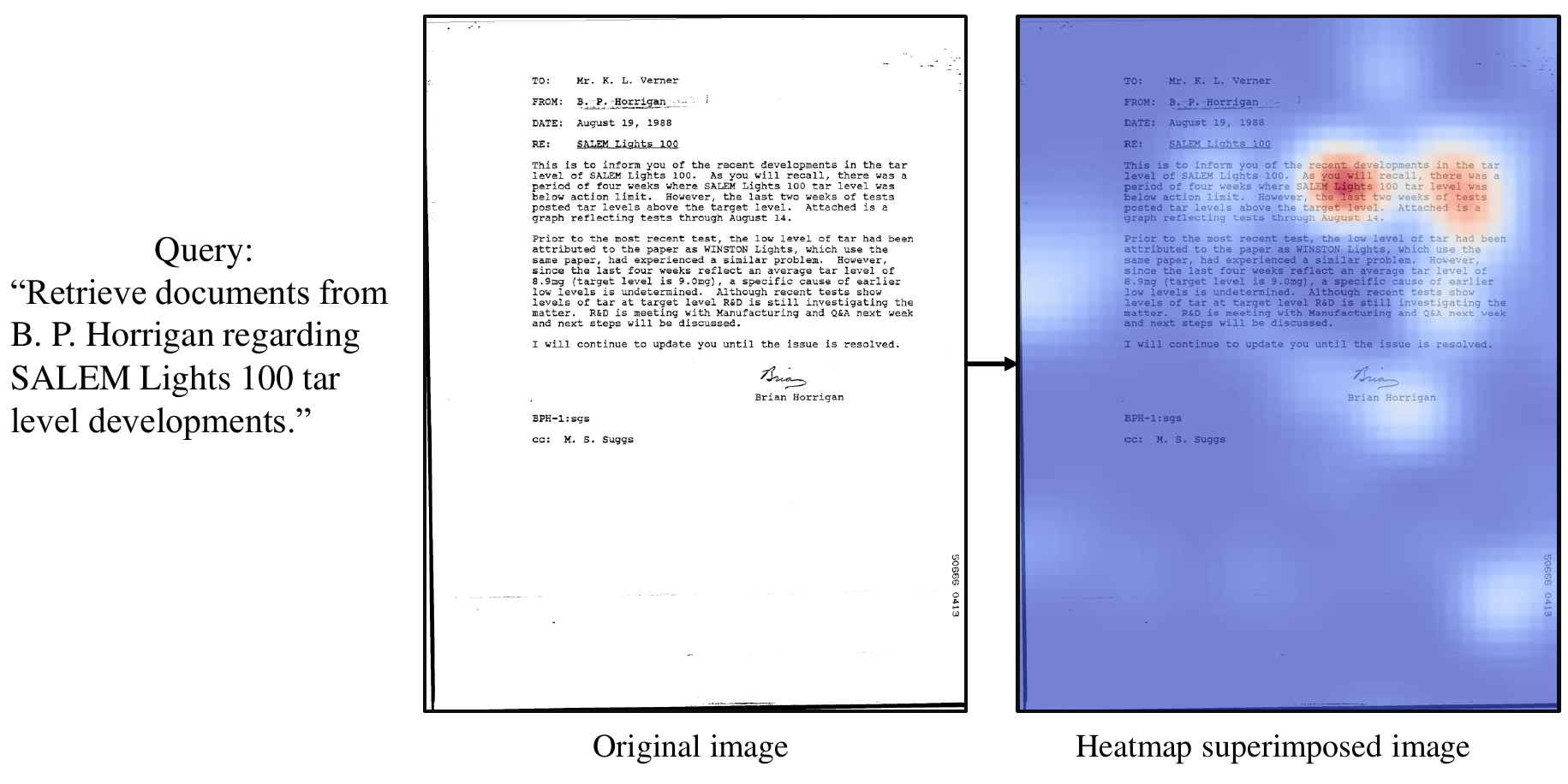}
\vspace{-5px}
\caption{Visualization of cross-attention in the re-ranking stage.} 
\label{fig:interpretability}
\vspace{-10px}
\end{figure}

The query corresponding to the image below is: “Retrieve documents from B. P. Horrigan regarding SALEM Lights 100 tar level developments.” As seen in \cref{fig:interpretability}, the areas related to "SALEM Lights 100 tar level developments" are prominently displayed. This indicates that the re-ranking stage has a certain degree of fine-grained matching and scoring capability, allowing for a more effective re-ranking of the original results.

\section{Additional Experimental Analysis}

To assess the generalization of models trained on our dataset to other tasks, we evaluate their performance on a downstream document classification task. As shown in Tab.\ref{tab:classification}, we compare the zero-shot classification results of the original and trained SigLIP models on the RVL-CDIP and Tobacco3482 datasets, demonstrating the effectiveness of our approach in improving document representation learning.

\begin{table}[!htbp]
\centering
\caption{Zero-shot comparison of original and trained Model.}
\vspace{-5px}
\resizebox{0.30\textwidth}{!}{
    \begin{tabular}{l|cc}
    \hline
    \textbf{Dataset} & \textbf{Original} & \textbf{Trained} \\ \toprule
        RVL-CDIP\cite{harley2015rvlcdip}       & 7.43 & 10.74 \\
        Tobacco3482\cite{kumar2012learning}   & 44.57 & 55.92 \\ \bottomrule
    \end{tabular}
}
\label{tab:classification}
\vspace{-10px}
\end{table}

Recent LVLMs, such as InternVL2-2B and Qwen2-VL-2B, have demonstrated strong document understanding capabilities. We leverage these models to generate content summaries for document images using the prompt: ``Please describe the document image.'' The generated descriptions serve as retrieval queries, which we then encode using the BGE model for document retrieval. As shown in Tab.\ref{tab:image-captioners}, the retrieval performance of these caption-based queries is comparable to that of OCR-IR. However, similar to OCR-IR, generating content summaries requires significant computational resources and time.

\begin{table}[!htbp]
\centering
\caption{Comparison with models as image-captioners ({CAP-IR}).}
\vspace{-5px}
\resizebox{0.48\textwidth}{!}{
\begin{tabular}{l|cc|c|c}
\hline
\textbf{Metric} & {\textbf{InternVL2-2B}} & {\textbf{Qwen2-VL-2B}}  & \textbf{OCR-IR} & \textbf{Ours} \\\toprule
    Recall@1      & 52.83  & 47.31  & 52.72  &81.03 \\ 
    Recall@10     & 71.63  & 68.07  & 72.16 &94.17  \\
    MRR@10    & 53.90  & 59.02  & 58.85  &85.68  \\ \bottomrule
\end{tabular}
}
\label{tab:image-captioners}
\vspace{-10px}
\end{table}

To gain deeper insights into the retrieval performance across different document categories, we further analyze the re-ranked retrieval results. As shown in Tab.\ref{tab:distribution}, we report the retrieval performance and the number of queries for five representative document categories.

\begin{table}[!htbp]
\centering
\caption{Retrieval performance on five representative categories.}
\vspace{-5px}
\resizebox{0.48\textwidth}{!}{
\begin{tabular}{l|ccccc}
    \toprule
    \textbf{Category}  & \textbf{Letter} & \textbf{Report} & \textbf{Memo} & \textbf{Form} & \textbf{Document} \\ \midrule
    \textbf{Query\_nums}  & 3675 & 1760 & 1520 & 1180 & 1175 \\
    Recall@1  & 83.10           & 82.67           & 87.50         & 73.64         & 89.62             \\ 
    Recall@10 & 93.88           & 95.80           & 94.61         & 94.66         & 97.96             \\ 
    MRR@10    & 86.86           & 87.37           & 89.92         & 81.33         & 92.98             \\ 
    \bottomrule
\end{tabular}
}
\label{tab:distribution}
\vspace{-10px}
\end{table}

\section{Future Work}
This study presents a preliminary exploration of document image retrieval, offering valuable insights into dataset construction and model optimization. However, as research progresses, several key directions warrant further investigation and improvement.

First, large-scale training data and the powerful representational capacity of LVLMs have enabled state-of-the-art retrieval performance in the recall stage. However, these models often incur significant computational and memory costs, raising concerns about efficiency. Currently, there is a lack of alignment models specifically designed for high-resolution document images and rich textual content. Effective cross-modal representation alignment facilitates the mapping of image and text information into a shared vector space, thereby enhancing fine-grained understanding and retrieval performance. This can help bridge the gap between cross-modal document image retrieval and purely text-based retrieval using OCR. Future research should focus on designing more efficient and compact models optimized for high-resolution document images and their textual content while improving image-text alignment. Furthermore, with advancements in generative models, the integration of cross-attention mechanisms with generative understanding models holds great potential. However, significant room remains for experimentation and improvements in the re-ranking stage. Despite the progress made in document image retrieval, a critical future direction lies in tightly integrating fine-grained generative understanding capabilities with the practical demands of document image retrieval.

Second, as document retrieval technology evolves, real-world applications often require retrieving multi-page documents. This necessitates models capable of processing and understanding multi-page document images while capturing long-range contextual dependencies. Additionally, there is a growing need for fine-grained paragraph-level retrieval. Currently, retrieval units in this study are typically single-page documents, and the models lack precise paragraph-level localization, which can impact retrieval accuracy in certain scenarios. Future research should explore long-document modeling for multi-page documents and precise paragraph-level localization. This is not only crucial for improving retrieval accuracy but also provides broader applications in document analysis and search systems.

In summary, future advancements in document image retrieval will focus on overcoming computational and memory efficiency bottlenecks, enhancing the ability to capture long-document information, and improving paragraph-level retrieval precision. As technology advances and application scenarios expand, document image retrieval is expected to play an increasingly vital role in improving information access efficiency and enhancing user experience.


\section{Prompt Design}
In this section, we provide the prompts used in the query generation and filtering processes, as shown in Table \cref{table:prompt}.

\begin{table*}[!htb]
\caption{The prompts used in the query generation and filtering processes.}
\resizebox{\textwidth}{!}{%
\centering
\small
\renewcommand\arraystretch{1.3}
\begin{tabular}{ll|l}
\toprule
\multicolumn{2}{c|}{\textbf{Model}} & \multicolumn{1}{c}{\textbf{Template}} \\ \midrule
\multicolumn{1}{l|}{\textbf{Generate}} & \textbf{ChatGPT \cite{john2023chatgpt}} & \begin{tabular}[c]{@{}l@{}}You are an expert who can use image's OCR to generate a query for retrieval. More \\ specifically, you will obtain the following content:\\ 1. Instruction: A statement used to describe specific task details. \\ 2. Layout-aware OCR Document: Text extracted from an image and arranged according to \\ to the layout when it appears in the image to maintain the relative position \\ between the texts appearing in the image. You need to understand the document layout\\  with the help of spaces and line breaks in the document.\\ NOW YOU TURN: \\ Instruction: You need to generate ten different layout-related queries that cover \\ all the different aspects of the entire document as much as possible. \\ These queries are used for retrieving layout-aware documents based on the above \\ conditions and the following. A query cannot be a simple and detailed description, \\ but should express the purpose of the search.\\ Layout-aware OCR Document : \textbf{\textcolor{blue}{\{document\}}} \\ Queries:\end{tabular} \\ \midrule
\multicolumn{1}{l|}{\multirow{2}{*}{\textbf{Score}}} & \textbf{ChatGPT \cite{john2023chatgpt}} & \begin{tabular}[c]{@{}l@{}}Here are ten queries used to retrieve image documents, and we would like to request\\  your feedback on the quality of the queries.\\ Please rate the quality of the ten given queries based on the content of the Layout\\ -aware OCR Document. Each query receives a score of 0 to 10, with higher scores \\ indicating higher quality.\\ Layout-aware OCR Document: \textbf{\textcolor{blue}{\{document\}}}\\ Queries: \textbf{\textcolor{blue}{\{queries\}}}\\ Please provide a comprehensive explanation of your evaluation to avoid any potential \\ biases.\\ Output format: \\ Scores: \\ Reasons:\end{tabular} \\ \cline{2-3} 
\multicolumn{1}{l|}{} & \textbf{Qwen-VL-Plus \cite{bai2023qwen}} & \begin{tabular}[c]{@{}l@{}}You are an expert in using images and their OCR text to score queries for retrieval.\\ Here are ten queries used to retrieve image documents, and we would like to request\\  your feedback on the quality of the queries.\\ Please rate the quality of the ten given queries based on the content of the Layout\\ -aware OCR Document and the document image. The document image, from which you can \\ obtain some visual elements that are not included in the Layout-aware OCR Document.\\ Each query receives a score of 0 to 10, with higher scores indicating higher quality. \\ Layout-aware OCR Document: \textbf{\textcolor{blue}{\{document\}}}\\ Queries: \textbf{\textcolor{blue}{\{queries\}}}\\ Please provide a comprehensive explanation of your evaluation to avoid any potential \\ biases.\\ Output format: \\ Scores: \\ Reasons:\end{tabular} \\ \bottomrule
\end{tabular}%
}
\label{table:prompt}
\end{table*}

\section{Licensing, Hosting and Maintenance Plan}\label{sec:lhm}

\paragraph{Author Statement.} 
We bear all responsibilities for the licensing, distribution, and maintenance of our dataset. 

\paragraph{License.} NL-DIR is under CC-BY-NC-SA 4.0 license. 

\paragraph{Hosting.} NL-DIR can be viewed and downloaded on huggingface at \url{https://huggingface.co/datasets/nianbing/NL-DIR}. Prior to the publication of the article, we typically present a selection of illustrative samples, after which we will release the entire dataset. We assure its long-term preservation for future reference and use. The annotations for retrieval queries are provided in the JSON file format, while the raw pictures are available in the PNG format. 

We do not hold any copyright for the document images; the copyrights belong to the UCSF Industry Documents Library and the document authors. For user convenience, we provide a download method for these document images, provided users agree that the data is only used for research purposes and not for commercial purposes. If copyright holders request the deletion or modification of certain images, we will hide or delete key information in the images to minimize the impact on the query. If the retention of images is not allowed, we will retain the query data and provide metadata for the corresponding images.

The Croissant metadata record is stored in \url{https://huggingface.co/datasets/nianbing/NL-DIR-sample/blob/main/croissant.json}. 

\paragraph{Metadata.} Metadata can be found at \url{https://huggingface.co/datasets/nianbing/NL-DIR}. 

\section{Datasheet}

\subsection{Motivation}
\textbf{For what purpose was the dataset created?} 

\textbf{\textcolor{red}{Answer: } }
NL-DIR establishes a fine-grained semantic retrieval dataset and benchmark for document images in real-world scenarios, which evaluates the retrieval performance of existing contrastive vision-language models (VLMs) and generative visual document understanding (VDU) models. NL-DIR provides an evaluation of existing models for document image understanding and cross-modal dense representation. As far as I know, NL-DIR is the first comprehensive benchmark for fine-grained document image semantic retrieval.

\subsection{Composition}
\textbf{What do the instances that comprise the dataset represent? (e.g., documents, photos, people, countries)} 

\textbf{\textcolor{red}{Answer: } }
Each instance represents a document image and five fine-grained semantic queries in our dataset. The document image is a PDF screenshot collected from UCSF Industry Documents Library \footnote{\url{https://www.industrydocuments.ucsf.edu}} in PNG format. The query is generated by LLM and then stored in a JSON file after being scored and manually filtered by a scoring model.

\textbf{How many instances are there in total (of each type, if appropriate)?} 

\textbf{\textcolor{red}{Answer: } }
We collected a total of 41,795 document images, each corresponding to five queries. The specific dataset statistics can be found in the main paper.

\textbf{Does the dataset contain all possible instances or is it a sample (not necessarily random) of instances from a larger set?} 

\textbf{\textcolor{red}{Answer: } }
We collect over 60K document images from the Industry Documents Library. The corresponding layout text information is extracted from the annotations of DocVQA \cite{mathew2021docvqa} and OCR-IDL \cite{ocridl2022}, which use Microsoft OCR and Amazon Textract respectively as OCR engines.

\textbf{Is there a label or target associated with each instance?} 

\textbf{\textcolor{red}{Answer: } }
Yes, for each document image, we generate and filter five queries.

\textbf{Is any information missing from individual instances?} 

\textbf{\textcolor{red}{Answer: } }
All instances are complete.

\textbf{Are relationships between individual instances made explicit (e.g., users’ movie ratings, social network links)?} 

\textbf{\textcolor{red}{Answer: } }
Some instances may have similar images or queries, but when filtering, we try to ensure a strong correlation between queries and images as much as possible.

\textbf{Are there recommended data splits (e.g., training, development/validation, testing)?}  

\textbf{\textcolor{red}{Answer: } }
Yes, we have done a reasonable split of the NL-DIR dataset, which is reflected in the already split JSON file, we will make all JSON files public after the publication.

\textbf{Are there any errors, sources of noise, or redundancies in the dataset?} 

\textbf{\textcolor{red}{Answer: } }
No.

\textbf{Is the dataset self-contained, or does it link to or otherwise rely on external resources (e.g., websites, tweets, other datasets)?}

\textbf{\textcolor{red}{Answer: } }
All data will be publicly accessible in the dataset repository. Our annotations will be stored in JSON format.

\textbf{Does the dataset contain data that might be considered confidential?} 

\textbf{\textcolor{red}{Answer: } }
No.

\textbf{Does the dataset contain data that, if viewed directly, might be offensive, insulting, threatening, or might otherwise cause anxiety?} 

\textbf{\textcolor{red}{Answer: } }
No.

\subsection{Collection Process}
The data collection process is described in the main paper and supplementary materials.

\subsection{Uses}
\textbf{Has the dataset been used for any tasks already?} 

\textbf{\textcolor{red}{Answer: } }
Yes, NL-DIR has been used to evaluate the cross-modal retrieval capabilities of as many as 9 different models.

\textbf{What (other) tasks could the dataset be used for?} 

\textbf{\textcolor{red}{Answer: } }
NL-DIR is mainly used for the evaluation of the cross-modal retrieval capability of document-related visual and language models.

\textbf{Is there a repository that links to any or all papers or systems that use the dataset?} 

\textbf{\textcolor{red}{Answer: } }
No.

\textbf{Is there anything about the composition of the dataset or the way it was collected and preprocessed/cleaned/labeled that might impact future uses?} 

\textbf{\textcolor{red}{Answer: } }
The document images we collected are all from IDL, and the corresponding OCR information is from DocVQA and OCR-IDL. The query generation and filtering methods have been provided in the main paper. However, we will do our best to maintain the dataset if the copyright holder requests the removal of certain data in the future.

\textbf{Are there tasks for which the dataset should not be used?} 

\textbf{\textcolor{red}{Answer: } }
No

\subsection{Distribution}
\textbf{Will the dataset be distributed to third parties outside of the entity (e.g., company, institution, organization) on behalf of which the dataset was created?} 

\textbf{\textcolor{red}{Answer: } }
Yes. The benchmark is publicly available on the Internet.

\textbf{How will the dataset will be distributed (e.g., tarball on website, API, GitHub)?} 

\textbf{\textcolor{red}{Answer: } }
The benchmark is available on Huggingface at \url{https://huggingface.co/datasets/nianbing/NL-DIR}.

\textbf{Will the dataset be distributed under a copyright or other intellectual property (IP) license, and/or under applicable terms of use (ToU)?} 

\textbf{\textcolor{red}{Answer: } }
CC-BY-NC-SA 4.0.

\textbf{Have any third parties imposed IP-based or other restrictions on the data associated with the instances?} 

\textbf{\textcolor{red}{Answer: } }
No.

\textbf{Do any export controls or other regulatory restrictions apply to the dataset or to individual instances?} 

\textbf{\textcolor{red}{Answer: } }
No.

\subsection{Maintenance}
\textbf{Who will be supporting/hosting/maintaining the dataset?} 

\textbf{\textcolor{red}{Answer: } }
The authors will be supporting, hosting, and maintaining the dataset.

\textbf{How can the owner/curator/manager of the dataset be contacted (e.g., email address)?} 

\textbf{\textcolor{red}{Answer: } }
Please contact the one of the authors (guohao2022@iie.ac.cn, qinxugong@njust.edu.cn).

\textbf{Is there an erratum?} 

\textbf{\textcolor{red}{Answer: } }
No. We will make announcements if there are any.

\textbf{Will the dataset be updated (e.g., to correct labeling errors, add new instances, delete instances)?} 

\textbf{\textcolor{red}{Answer: } }
Yes. We will post a new update in \url{https://huggingface.co/datasets/nianbing/NL-DIR} if there is any.

\textbf{If the dataset relates to people, are there applicable limits on the retention of the data associated with the instances (e.g., were individuals in question told that their data would be retained for a fixed period and then deleted)?} 

\textbf{\textcolor{red}{Answer: } }
People's information may appear in the reference images. People may contact us to exclude specific data instances if they appear in the reference images.

\textbf{Will older versions of the dataset continue to be supported/hosted/maintained?} 

\textbf{\textcolor{red}{Answer: } }
Yes. Old versions will also be hosted in \url{https://huggingface.co/datasets/nianbing/NL-DIR-sample}

\textbf{If others want to extend/augment/build on/contribute to the dataset, is there a mechanism for them to do so?} 

\textbf{\textcolor{red}{Answer: } }
Yes, according to our dataset construction method, if the data is compliant and reasonable, expanding the dataset is allowed.

\clearpage

\end{document}